\pdfoutput=1
\documentclass{article}
\usepackage{arxiv}
\usepackage[utf8]{inputenc} 
\usepackage[T1]{fontenc}    
\usepackage{hyperref}       
\usepackage{url}            
\usepackage{booktabs}       
\usepackage{amsfonts}       
\usepackage{nicefrac}       
\usepackage{microtype}      
\usepackage{lipsum}		
\usepackage{graphicx}
\usepackage{doi}
\usepackage[dvipsnames,table,xcdraw]{xcolor}  
\usepackage{amsmath}  
\usepackage{longtable}
\usepackage{multirow}
\usepackage{makecell}
\usepackage{multicol}
\usepackage{subfigure}
\usepackage{caption}
\usepackage[para,online,flushleft]{threeparttable}
\usepackage{tablefootnote}
\title{A survey of emerging applications of diffusion probabilistic models in MRI}

\author{ 
Yuheng Fan\thanks{This work was done as a visiting student at the School of Biomedical Engineering, ShanghaiTech University.}\\
School of Biomedical Engineering\\
ShanghaiTech University\\
Pudong, Shanghai, China, 201210 \\
\texttt{yuhengfan@outlook.com} \\
\And
Hanxi Liao\\
School of Biomedical Engineering\\
ShanghaiTech University\\
Pudong, Shanghai, China, 201210 \\
\texttt{liaohx2023@shanghaitech.edu.cn} \\
\And
Shiqi Huang\\
School of Biomedical Engineering\\
ShanghaiTech University\\
Pudong, Shanghai, China, 201210\\
\texttt{huangshq1@shanghaitech.edu.cn} \\
\And
Yimin Luo\\
School of Biomedical Engineering and Imaging Sciences\\
King’s College London\\
St. Thomas’ hospital, London, SE1 7EH, U.K. \\
\texttt{yimin.luo@kcl.ac.uk} \\
\And
Huazhu Fu\\
Institute of High Performance Computing\\
Agency for Science, Technology\\ and Research (A*STAR), Singapore, 201210\\
\texttt{hzfu@ieee.org} \\
\And
Haikun Qi\thanks{Corresponding author.}\\
School of Biomedical Engineering\\
ShanghaiTech University\\
Pudong, Shanghai, China, 201210\\
\texttt{qihk@shanghaitech.edu.cn} \\
}

\date{}

\renewcommand{\shorttitle}{A survey of emerging applications of diffusion probabilistic models in MRI}

\hypersetup{
pdftitle={A survey of emerging applications of diffusion probabilistic models in MRI},
pdfsubject={cs.LG,},
pdfauthor={Yuheng Fan, Hanxi Liao, Shiqi Huang, Yimin Luo, Huazhu Fu, Haikun Qi},
pdfkeywords={Diffusion Probabilistic Models, Score-Based Generative Modeling, MRI},
}

\begin{document}
\maketitle

\begin{abstract}
Diffusion probabilistic models (DPMs) which employ explicit likelihood characterization and a gradual sampling process to synthesize data, have gained increasing research interest. Despite their huge computational burdens due to the large number of steps involved during sampling, DPMs are widely appreciated in various medical imaging tasks for their high-quality and diversity of generation. Magnetic resonance imaging (MRI) is an important medical imaging modality with excellent soft tissue contrast and superb spatial resolution, which possesses unique opportunities for DPMs. Although there is a recent surge of studies exploring DPMs in MRI, a survey paper of DPMs specifically designed for MRI applications is still lacking. This review article aims to help researchers in the MRI community to grasp the advances of DPMs in different applications. We first introduce the theory of two dominant kinds of DPMs, categorized according to whether the diffusion time step is discrete or continuous, and then provide a comprehensive review of emerging DPMs in MRI, including reconstruction, image generation, image translation, segmentation, anomaly detection, and further research topics. Finally, we discuss the general limitations as well as limitations specific to the MRI tasks of DPMs and point out potential areas that are worth further exploration. 
\end{abstract}


\section{Introduction}
Magnetic Resonance Imaging (MRI), with superior soft tissue contrast, noninvasive nature, and multiplanar imaging capability, has become an important imaging modality in medical diagnosis and therapy. However, it is also constrained by complex imaging principles, long scan times, and high economic costs, which may hinder its full potential for clinical applications. 

In recent years, the flourishing development of generative artificial intelligence (AI) models has provided new and promising solutions to tackle the challenges associated with MRI. For example, works in \cite{volokitin2020modelling,edupuganti2020uncertainty} applied variational auto-encoder (VAE) in MRI synthesis and reconstruction. Some works \cite{han2021madgan, zhao2020tripartite, chen2020mri} proposed to design models for anomaly detection, reconstruction, and super-resolution of MR images based on Generative Adversarial Networks (GAN) \cite{goodfellow2020generative}. These models lead to high-quality image generation, data-driven optimization approaches, and potential cost-effective application integration. However, VAE \cite{kingma2013auto} is limited by its formation of the posterior and prior distributions, requiring a trade-off between structural complexity and model representational capacity, and tends to generate low-fidelity samples. Normalization Flow \cite{rezende2015variational} enables accurate computation of the probability likelihood function, but it is constrained by the requirement that each transformation of the model be reversible, which restricts the design of the model. Auto-regressive Models \cite{bengio2000taking,larochelle2011neural}, as another classical generative models, also face issues such as slow generation speed with large decoding space and difficulties in generating high-resolution images. In recent years, GAN has shown outstanding generative performance, but its adversarial learning mechanism  causes training instability.

\begin{figure*}[htbp]
        \centering
        
	\subfigure[]{
		\begin{minipage}[b]{0.4\linewidth}
			\includegraphics[width=1\linewidth]{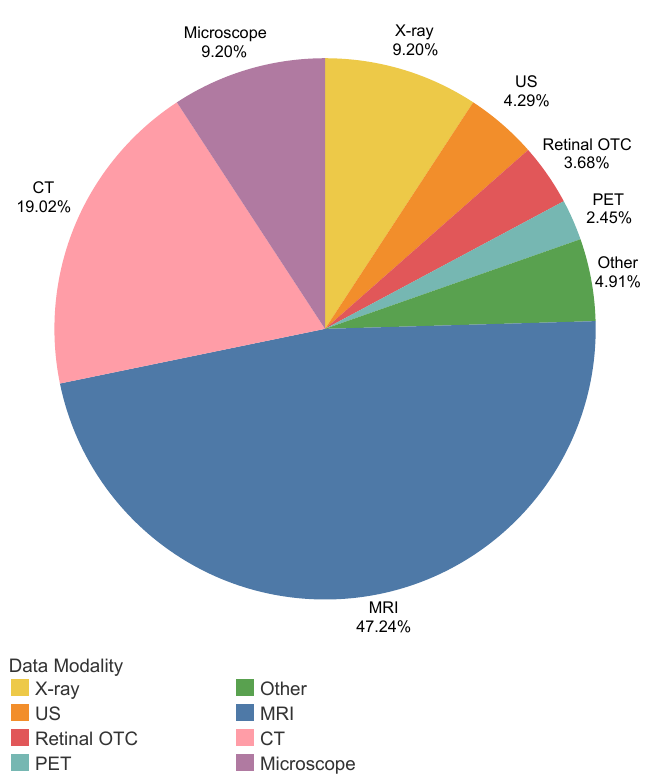}
		\end{minipage}
            
	}
    	\subfigure[]{
    		\begin{minipage}[b]{0.4\linewidth}
   		 	\includegraphics[width=1\linewidth]{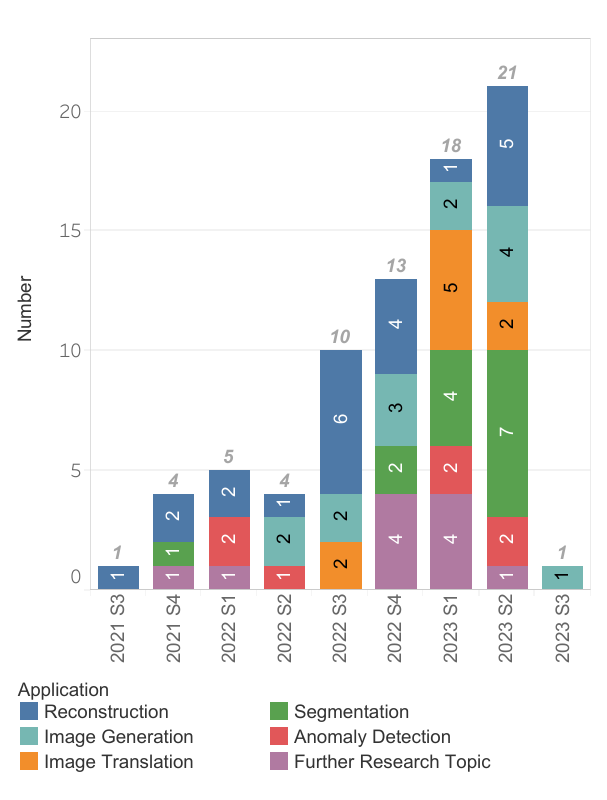}
    		\end{minipage}
    	}
	\caption{Development of Diffusion Probabilistic Models in Medical Imaging and Emerging Application in MRI. (a) Pie chart of the medical imaging modality to which DPMs have been applied. (b) Histogram of DPMs in different applications of MRI.}
\label{fig:Sec1-PieHist}
\end{figure*}

Diffusion Probabilistic Models (DPMs), as a newly emerging family of generative models, have attracted considerable attention in the field of medical imaging due to their well-established mathematical explanations, adversarial-free training strategy, and ability to achieve stable and controllable generation. By collecting all the methods of applying DPMs in medical imaging that have emerged from 2021 to the third quarter of 2023 and analyzing the relevant data modalities in Fig. \ref{fig:Sec1-PieHist}(a), we found that about 47.24\% of the methods focus on MRI. Furthermore, the studies applying DPMs in MRI over the years as summarized in Fig. \ref{fig:Sec1-PieHist}(b), indicate that the application of DPMs in MRI has shown a rapid development trend of expanding scope and increasing quantity. This trend is attributed to the unique advantages of DPMs in meeting MRI needs. Indeed, as a versatile diagnostic tool, MRI can produce rich contrasts for comprehensive diagnosis, while it also faces some long-standing challenges, such as low acquisition speed and being vulnerable to motion. Besides DPMs' capacity to solve the inverse problem of MRI reconstruction\cite{song2021solving}, benefiting from their unique setting of noise estimation, the reconstruction process is also accompanied by the denoising of the image, which can achieve accelerated reconstruction and noise reduction simultaneously. Furthermore, the better mathematical interpretation of conditional DPMs provides a more stable and reliable method for image synthesis, which can be used to translate one or more MRI contrasts to a target contrast or translate between MRI and other imaging modalities. Also, the excellent generation performance makes it a new tool for understanding functional MRI\cite{chen2023seeing, takagi2023high}. Therefore, a comprehensive review and in-depth analysis of the emerging application of DPMs in MRI is of great importance.

We hope that this paper can serve as a good starting point for researchers in the MRI community interested in this fast-developing and important field. The main contributions of this paper lie in the following aspects:

\begin{itemize}
    \item \textbf{A holistic overview of the fundamentals of DPMs.} We summarize the principles of two currently dominant classes of DPMs from the perspective of the formation of the diffusion time step, revealing the relationship between the two classes of models, and then elucidate conditional DPMs.
    
    \item \textbf{A systematical survey on the applications of DPMs in MRI.} We describe in detail the studies of applying DPMs to different tasks in MRI, including the well-known topics of image reconstruction, image generation and translation, segmentation, and anomaly detection, as well as other pioneering research topics such as registration, motion correction, super-resolution, and additional emerging downstream applications.
    
    \item \textbf{An in-depth discussion of trends and challenges.} We discuss in depth the trends and challenges of applying DPMs to MRI, revealing future directions of DPM developments, including model design and expanding applications.
\end{itemize}

\section{Theory}\label{Sec:Theory}

\begin{figure}[htbp]
	\centering
    \includegraphics[width=0.95\linewidth]{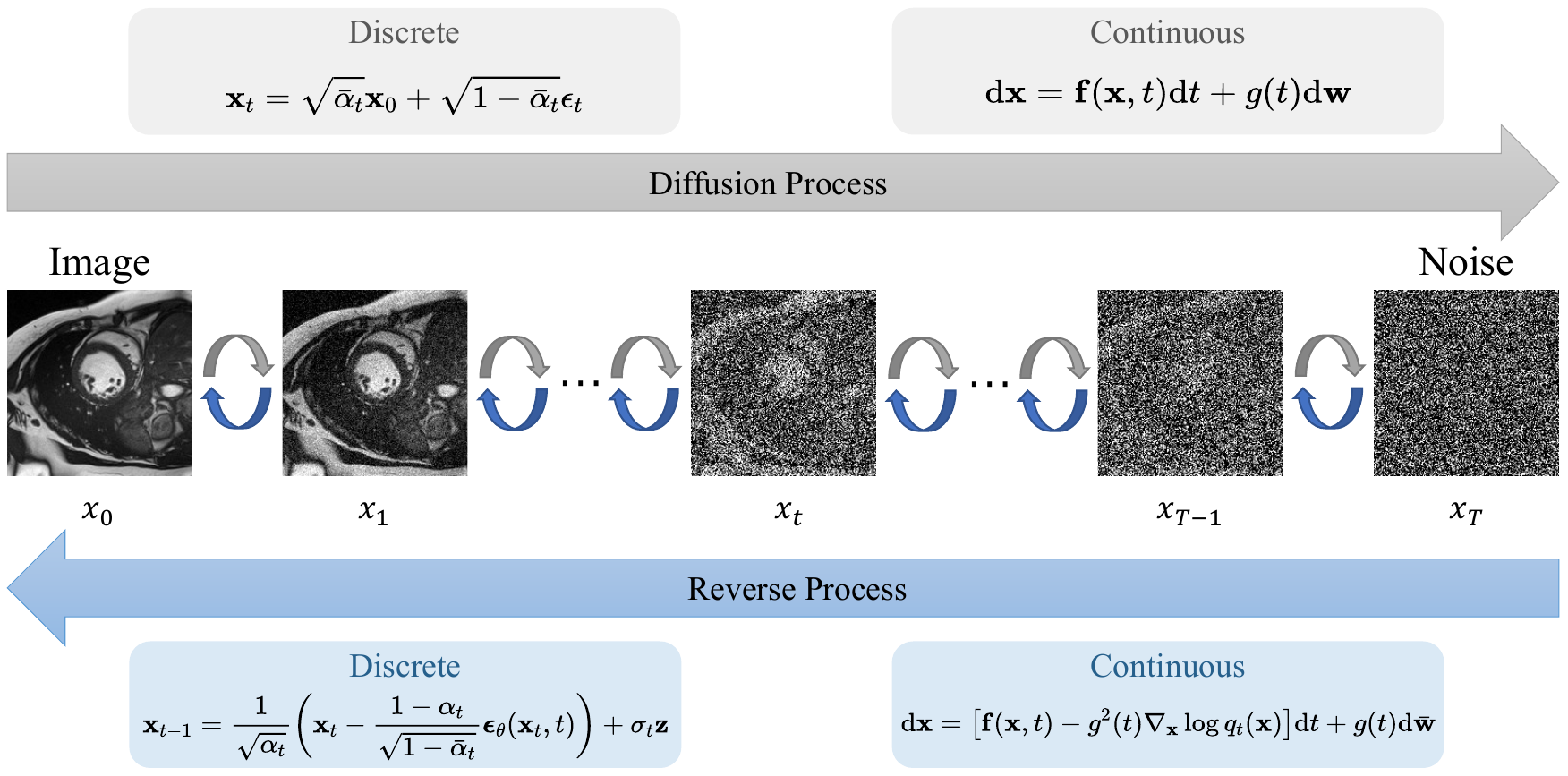}
    \captionsetup{justification=centering}
	\caption{Diffusion Process and Reverse Process in Diffusion Probabilistic Models (DPMs).}
 \label{fig:two-process}
\end{figure}

Diffusion Probabilistic Models (DPMs), a new paradigm for generative models, are proposed to use a neural network to estimate the representation of a series of Gaussian noises $\epsilon_t \sim \mathcal{N}(\mu_t,\sigma_t)$ that could perturb data $x$ into noise with standard normal distribution $z \sim \mathcal{N}(\mathbf{0}, \mathbf{I})$ (called the "diffusion process"), and then gradually recover the data sample $\hat{x}$ from $z$ (called the "reverse process"). Fig. \ref{fig:two-process} demonstrates the diffusion and the reverse process.

Estimation of the representations under different diffusion time steps leads to two types of Diffusion Probabilistic Models: discrete-time diffusion models involving noise estimation and Markov Chain and continuous-time diffusion models involving score matching and Stochastic Differential Equations (SDEs). This section will introduce the principles of the two main aspects of DPMs: (1) the \textit{diffusion process}; (2) the \textit{reverse process} and its corresponding training objective. 

\subsection{Discrete-time Diffusion Models}
\subsubsection{DDPM}
Sohl-Dickstein et al. \cite{sohl2015deep} firstly introduced the principle of DPMs, which converts a simple known distribution into a target distribution using a generative Markov chain. 

For a given data distribution $q(\mathbf{x}_0)$, the \textit{diffusion process} is characterized by a discrete-time Markov chain $\{\mathbf{x}_t, 0 \le t \le T, t\in \mathbb{N} \}$ with transition probability $q(\mathbf{x}_t |\mathbf{x}_{t-1} )$, and according to the Markov property, the relationship between $q(\mathbf{x}_0)$ and the stationary distribution $p(\mathbf{x}_T) \sim \mathcal{N}(\mathbf{0},\mathbf{I})$ of the Markov chain is given by Eq. \ref{DDPMforward-prob}.
\begin{align}
\begin{aligned}
\label{DDPMforward-prob}
&q\left(\mathbf{x}_{1},\dots, \mathbf{x}_{T} \mid \mathbf{x}_0\right)=\prod_{t=1}^T q\left(\mathbf{x}_t \mid \mathbf{x}_{t-1}\right) \\ &q\left(\mathbf{x}_t \mid \mathbf{x}_{t-1}\right)=\mathcal{N}\left(\mathbf{x}_t ; \sqrt{1-\beta_t} \mathbf{x}_{t-1}, \beta_t \mathbf{I}\right)
\end{aligned}
\end{align}
where the noise schedule $\beta_t \in (0,1)$ is a set of hyperparameters that are usually set linearly increased, reflecting the level of the noise added to the original signal at each transition. 

With the notation of $\alpha_t = 1-\beta_t$, $\bar{\alpha}_t=\Pi_{i=1}^{t}\alpha_i$, and the transition probability $q(\mathbf{x}_t |\mathbf{x}_{t-1} )$, we could use Eq. \ref{DDPM-forward-trans} to transform a given sample $\mathbf{x}_0 \sim p(\mathbf{x}_0)$ into a noisy data $\mathbf{x}_t$ with noise $\epsilon_t \sim \mathcal{N}(\mathbf{0},\mathbf{I})$.

\begin{equation}
\label{DDPM-forward-trans}
\mathbf{x}_t = \sqrt{\bar{\alpha}_t} \mathbf{x}_0 + \sqrt{1-\bar{\alpha}_t}\epsilon_t
\end{equation}

For the \textit{reverse process}, the transition of the reverse-time Markov chain is approximated by the learnable conditional probability $p_\theta(\mathbf{x}_{t-1}|\mathbf{x}_{t})$ with the setting of Eq. \ref{DDPM-reverse-prob}.


\begin{equation}
\label{DDPM-reverse-prob}
p_\theta(\mathbf{x}_{t-1}|\mathbf{x}_{t}) = \mathcal{N}(\mathbf{x}_{t-1};\mu_\theta(\mathbf{x}_t,t),\Sigma_{\theta}(\mathbf{x}_t,t))
\end{equation}
where the mean $\mu_\theta(\mathbf{x}_t,t)$ and variance $\Sigma_{\theta}(\mathbf{x}_t,t)=\sigma_t^2\mathbf{I}$ are learned by a neural nework $\epsilon_\theta(\mathbf{x}_t, t)$ with $\theta$ denoting network parameters, and $\sigma_t$ is commonly set to a fixed $\beta_t$ or $\tilde{\beta}_t = \frac{1-\bar{\alpha}_{t-1}}{1-\bar{\alpha}_t}\beta_{t}$. Thus, we could first sample from $\mathbf{x}_T \sim p(\mathbf{x}_T)$, and then iteratively sample $x_{t-1}$ according to the learned $p_\theta(\mathbf{x}_{t-1}|\mathbf{x}_{t})$ until $t=1$, to get the generated result $\hat{\mathbf{x}}_0 \sim p(\mathbf{x}_0) $. Such a sampling process can be described as Eq. \ref{DDPM-reverse-trans}. 
\begin{align}
    \begin{aligned}
        \label{DDPM-reverse-trans}
\mathbf{x}_{t-1}&=\frac{1}{\sqrt{\alpha_t}}\left(\mathbf{x}_t-\frac{1-\alpha_t}{\sqrt{1-\bar{\alpha}_t}} \boldsymbol{\epsilon}_\theta\left(\mathbf{x}_t, t\right)\right)+\sigma_t \mathbf{z}\\&\text{where }\mathbf{z}\sim \mathcal{N}(\mathbf{0},\mathbf{I}) \mbox{ if } t>0 \mbox{ else } \mathbf{z}=\mathbf{0}
    \end{aligned}
\end{align}

For the network optimization, Sohl-Dickstein et al. \cite{sohl2015deep} and Ho et al. \cite{ho2020denoising} indicated that we could derive a simplified optimization objective via minimizing the variational bound on the negative log-likelihood:

\begin{align}
    \begin{aligned}
    \label{DDPM-obj}
        L(\theta) = \mathbb{E}_{\mathbf{x}_0, \boldsymbol{\epsilon}}\left[\frac{\beta_t^2}{2 \sigma_t^2 \alpha_t\left(1-\bar{\alpha}_t\right)}\cdot\left\|\boldsymbol{\epsilon}-\boldsymbol{\epsilon}_\theta\left(\sqrt{\bar{\alpha}_t} \mathbf{x}_0+\sqrt{1-\bar{\alpha}_t} \boldsymbol{\epsilon}, t\right)\right\|^2
        \right]
    \end{aligned}
\end{align}

And Ho et al. \cite{ho2020denoising} pointed out that Eq. \ref{DDPM-obj} can be reduced to Eq. \ref{DDPM-obj-simple}, where $t$ is uniform between $1$ and $T$. By comparing Eq. \ref{DDPM-obj} to Eq. \ref{DDPM-obj-simple}, we can see that the weighting of the two noises in Eq. \ref{DDPM-obj-simple} is ignored in Eq. \ref{DDPM-obj}. Apart from simplifying the optimization objective, this simplification can also reduce the weighting of the estimation for small amounts of noisy perturbation, thus allowing the network to focus on the more difficult estimation when large amounts of noise are added to the image. Better experimental results of Eq. \ref{DDPM-obj-simple} used more frequently in practice.

\begin{equation}
\label{DDPM-obj-simple}
L_{\text{simple}}(\theta) = \mathbb{E}_{t, \mathbf{x}_0, \boldsymbol{\epsilon}}\left[\left\|\boldsymbol{\epsilon}-\boldsymbol{\epsilon}_\theta\left(\sqrt{\bar{\alpha}_t} \mathbf{x}_0+\sqrt{1-\bar{\alpha}_t} \boldsymbol{\epsilon}, t\right)\right\|^2\right]
\end{equation}

In fact, $\epsilon_\theta$ estimates the noise added to  $\mathbf{x}_t$ in the \textit{diffusion process} described by Eq. \ref{DDPM-forward-trans}. Therefore, Eq. \ref{DDPM-reverse-trans} is a denoising model, and the \textit{reverse process} represents the removal of Gaussian noise to obtain a clean image.

\subsubsection{DDIM}

One major limitation of DDPM \cite{ho2020denoising} is its long generation time, which arises from the need to simulate denoised results along the Markov Chain iteratively. Different from DDPM using Markov property, the Denoising Diffusion Implicit Models (DDIM) \cite{song2020denoising} achieved sampling acceleration by directly defining the $q(\mathbf{x}_{t-1}|\mathbf{x}_t, \mathbf{x}_0)$ that does not rely on the \textit{diffusion process}, so it is possible to reduce the number of iterations by crossing part of the intermediate state in the \textit{reverse process}. More importantly, since the generation is deterministic when $\mathbf{x}_T$ is fixed, multiple samples conditioned on one latent variable should have similar high-level features, which constitutes the basis of conditional diffusion probabilistic models.

The \textit{diffusion process} and training objective of DDIM are similar to DDPM. However, for the \textit{reverse process}, DDIM proposed to replace $q(\mathbf{x}_{t-1}|\mathbf{x}_t, \mathbf{x}_0)$ with $   q_\sigma(\mathbf{x}_1, \dots, \mathbf{x}_T | \mathbf{x}_0) = q_{\sigma}(\mathbf{x}_T | \mathbf{x}_0) \prod_{t=2}^Tq_{\sigma}(\mathbf{x}_{t-1} | \mathbf{x}_t, \mathbf{x}_0)$, where $\sigma \in \mathbb{R}_{\ge0}^{T}$ is an index of the generated distribution related to the \textit{reverse process} and

\begin{equation}
    q_{\sigma}(\mathbf{x}_T|\mathbf{x}_0)  = \mathcal{N}(\sqrt{\bar{\alpha}_T}\mathbf{x}_0, (1-\bar{\alpha}_T ) \mathbf{I} )
\end{equation}
For $1 < t < T$, $q_{\sigma}(\mathbf{x}_{t-1}|\mathbf{x}_t,\mathbf{x}_0)$ satisfies the distribution in Eq. \ref{DDIM-define-prob}. The motivation of this setup is to ensure that the joint distribution defined as Eq. \ref{DDIM-define-prob} still match the marginals $q_{\sigma}(\mathbf{x}_t|\mathbf{x}_0)  = \mathcal{N}(\sqrt{\bar{\alpha}_t}\mathbf{x}_0, (1-\bar{\alpha}_t ) \mathbf{I} )$ 
for all $1 < t < T$. 
\begin{align}
\label{DDIM-define-prob}
\mathcal{N}(\sqrt{\bar{\alpha}_{t-1}}\mathbf{x}_0 + \sqrt{1-\bar{\alpha}_{t-1}-\sigma ^2_{t}} \cdot \frac{\mathbf{x}_t - \sqrt{\bar{\alpha}_t}\mathbf{x}_0}{\sqrt{1-\bar{\alpha}_{t}}}, \sigma_{t}^2 \mathbf{I} )  
\end{align}

Since the corresponding "forward process" means that every $\mathbf{x}_t$ depends on $\mathbf{x}_{t-1}$ and $\mathbf{x}_0$, this process is non-Markovian. Specifically, for the \textit{reverse process}, the approximation of $p_\theta(\mathbf{x}_{t-1}|\mathbf{x}_{t})$ can start with a standard normal distribution $p_\theta (\mathbf{x}_T)= \mathcal{N}(\mathbf{0}, \mathbf{I})$ when $t = T$. During the intermediate state $1<t<T$, we first get $\mathbf{x}_0= (\mathbf{x}_t - \sqrt{1-\bar{\alpha}_t}\epsilon_t)/ \sqrt{\bar{\alpha_t}}$ by Eq. \ref{DDPM-forward-trans} for a given $\mathbf{x}_t$. Next, by the noise estimation $ \epsilon_\theta(\mathbf{x}_t, t)\approx \epsilon_t$, we can approximate $p_\theta(\mathbf{x}_{t-1}|\mathbf{x}_{t})$ by replacing $\mathbf{x}_0$ with $(\mathbf{x}_t - \sqrt{1-\bar{\alpha}_t}\cdot \epsilon_\theta(\mathbf{x}_t, t))/ \sqrt{\bar{\alpha_t}}$ in $q_{\sigma}(\mathbf{x}_{t-1}|\mathbf{x}_t, \mathbf{x}_0)$, where $q_{\sigma}(\mathbf{x}_{t-1}|\mathbf{x}_t, \mathbf{x}_0)$ is defined as Eq. \ref{DDIM-define-prob}. This replacement is shown in Eq. \ref{reverse-estimation-prob}. Finally, we can predict the denoised observation $\hat{\mathbf{x}}_0$ via $p_\theta(\mathbf{x}_{0}|\mathbf{x}_{1}) \approx \mathcal{N}((\mathbf{x}_1 - \sqrt{1-\bar{\alpha}_1} \cdot \epsilon_{\theta} (\mathbf{x}_1, 1 ))/\sqrt{\bar{\alpha}_1},\sigma_1^2\mathbf{I})$ when $t =1$.

\begin{align}
\label{reverse-estimation-prob}
    p_\theta(\mathbf{x}_{t-1}|\mathbf{x}_{t})\approx q_{\sigma}\left(\mathbf{x}_{t-1}|\mathbf{x}_t, \frac{\mathbf{x}_t - \sqrt{1-\bar{\alpha}_t} \cdot \epsilon_{\theta} (\mathbf{x}_t, t )}{\sqrt{\bar{\alpha}_t}}\right)
\end{align}

According to the approximation of $p_\theta(\mathbf{x}_{t-1}|\mathbf{x}_{t})$ as explained above, the sampling process from $\mathbf{x}_t$ to $\mathbf{x}_{t-1}$ can be described as Eq. \ref{reverse-estimation-trans} by using the reparameterization trick to modify the replaced $q_{\sigma}(\mathbf{x}_{t-1}|\mathbf{x}_t, \mathbf{x}_0)$ defined as Eq. \ref{DDIM-define-prob}.

\begin{align}
\begin{aligned}
\label{reverse-estimation-trans}
\mathbf{x}_{t-1}=\sqrt{\bar{\alpha}_{t-1}}\underbrace{\left(\frac{\mathbf{x}_{t}-\sqrt{1-\bar{\alpha}_{t}} \cdot \epsilon_\theta(\mathbf{x}_t, t)}{\sqrt{\bar{\alpha}_{t}}}\right)}_{\text {"predicted } \mathbf{x}_{0} \text { " }} +\underbrace{\sqrt{1-\bar{\alpha}_{t-1}-\sigma_{t}^{2}} \cdot \epsilon_\theta(\mathbf{x}_t, t)}_{\text {"direction pointing to } \mathbf{x}_{t} \text { " }}+\underbrace{\sigma_{t} \epsilon_{t}}_{\text {random noise }}
\end{aligned}
\end{align}

\noindent where $\epsilon_t \sim \mathcal{N}(\mathbf{0}, \mathbf{I})$ is a standard Gaussian noise independent of $\mathbf{x}_t$,  $\epsilon_\theta(\mathbf{x}_t, t)$ is the noise estimation provided by the network. Intuitively, the first term of the summation on the right-hand side comes directly from the representation of $\mathbf{x}_0$ given $\mathbf{x}_t$ and thus denotes the predicted $\mathbf{x}_0$, while the second term of the summation comes from the correction of the mean function that matches the marginals of $\mathbf{x}_t$ in Eq. 8 and thus denotes the direction pointing to $\mathbf{x}_t$. Moreover, Song et al. \cite{song2020denoising} pointed out that different choices of $\sigma$ result in different sampling processes. Specifically, let $\sigma_t^{2} = \eta\cdot\tilde{\beta}_t$ and $\eta \ge 0$  is a hyperparameter related to the noise intensity. If $\eta = 1$, the sampling process is equivalent to DDPM, while if $\eta = 0$, the generation is free of random noise and becomes deterministic when the original $\mathbf{x}_T$ has been generated. Moreover, accelerated sampling could be achieved by replacing the sequence $[1,\dots,T]$ with its subsequence $[\tau_1,\dots,\tau_S], \ S\le T$ in Eq. \ref{reverse-estimation-trans}.

\subsection{Continuous-time Diffusion Models}

The target of Diffusion Probabilistic Models with the two processes in Fig. \ref{fig:two-process} is to find a stable iterative modeling of the data distribution $q(\mathbf{x})$. When $t$ in $\{\mathbf{x}_t\}$ changes from a discrete-time $0 \le t \le T, t\in \mathbb{N}$ to a continuous scenario $t\in[0, T]$,  $\{\mathbf{x}_t\}$ is no longer a discrete-time Markov chain but a stochastic process with Markov properties (i.e., the Markov process), which provides a new mathematical tool into Diffusion Probabilistic Models.

\subsubsection{Score matching with Langevin dynamics}

Unlike DDPM using the Markov chain, Score matching with Langevin dynamics (SMLD) \cite{song2019SMLD} approximate the probability density $q(\mathbf{x})$ by estimating the gradient of log-likelihood $\nabla_\mathbf{x} \log q(\mathbf{x})$ (i.e. the (Stein) score function \cite{score-function}) at each noise scale with the neural network $s_\theta(\mathbf{x}),s_{\theta}:\mathbb{R}^D\to\mathbb{R}^D$. Therefore, SMLD can make the probability density of the Energy-based Models \cite{lecun2006tutorial} computable by replacing its normalization constant.

\begin{equation}
\label{score-matching}
\arg \min_{\theta}\mathbb{E}_{q(\mathbf{x})}\left[||\nabla_\mathbf{x} \log q(\mathbf{x})-s_{\theta}(\mathbf{x})||^2_2\right]
\end{equation}

The objective of approximating the score $\nabla_\mathbf{x} \log q(\mathbf{x})$ using $s_\theta(\mathbf{x})$ can be described as minimizing Fisher divergence, as in Eq. \ref{score-matching}, and studies about score matching  \cite{scorematching-1,scorematching-2,scorematching-3} provided methods for minimizing the Fisher divergence on the training set when $\nabla_\mathbf{x} \log q(\mathbf{x})$ is unknown.

However, under the manifold hypothesis, the estimation of the score function in the low-density region will be inaccurate due to the small number of data points for score matching. Particularly, the low-density portion of $q(\mathbf{x})$ is neglected in the integration of Eq. \ref{expand-score-matching} when $q(\mathbf{x})$ is relatively small.

\begin{align}
\begin{aligned}
\label{expand-score-matching}
\mathbb{E}_{q(\mathbf{x})}\left[||\nabla_\mathbf{x} \log q(\mathbf{x})-s_{\theta}(\mathbf{x})||^2_2\right] = \int q(x) ||\nabla_\mathbf{x} \log q(\mathbf{x})-s_{\theta}(\mathbf{x})||^2_2 \mathrm{d}\mathbf{x}
\end{aligned}
\end{align}

Therefore, the author in \cite{song2019SMLD} proposed adding Gaussian noise $\epsilon_t$ of different intensities to the data distribution as Eq. \ref{Langevin-Dynamic}. Thus, the score function $\nabla_{\mathbf{x}} \log q(\mathbf{x})$ can cover the space uniformly by each estimated $\nabla_{\mathbf{x}_t} \log q(\mathbf{x}_t)$, making the training of the score estimator $s_\theta(\mathbf{x}_t)$ more stable.

\begin{equation}
\label{Langevin-Dynamic}
\mathbf{x}_t = \mathbf{x}_{t-1} + \frac{\delta}{2} \nabla_{\mathbf{x}_{t-1}} \log p(\mathbf{x}_{t-1} ) + \sqrt{\delta} \boldsymbol{\epsilon}_t
,\
\boldsymbol{\epsilon}_t \sim \mathcal{N}(\mathbf{0}, \mathbf{I})
\end{equation}

Where $\delta$ is the step size and the error in estimating $q(\mathbf{x})$ using $p(\mathbf{x}_t)$ can be sufficiently small when $t \to T$ and $T\to \infty$ under a small step size $\delta \to 0$. Subsequently, we can use the Stochastic Gradient Langevin Dynamic \cite{LangevinDynamic} in Eq. \ref{Langevin-Dynamic} to obtain $q(\mathbf{x})$ using $s_{\theta}(\mathbf{x}_{t-1}) \approx \nabla_{\mathbf{x}_{t-1}} \log q(\mathbf{x}_{t-1})$. 

\subsubsection{Score-based SDE} \label{sec:Score-based}

Score-based SDE \cite{song2020score} innovatively examined the DPMs from the perspective of SDE. It proposed that the \textit{diffusion process} and the \textit{reverse process} has its corresponding SDE, and that generating samples in the \textit{reverse process} is equivalent to utilizing $s_\theta$ to get the numerical solution of the reverse-SDE. This work also proved that for all diffusion processes, there exists a deterministic process described by the ordinary differential equation (ODE) Despite the theoretical similarities between DDPM and SMLD demonstrated by this observation, it also provides a unifying perspective on the theoretical study of DPMs.

Specifically, since the Markov process $\{\mathbf{x}_t, t\in[0, T]\}$  has continuous sampling paths, Score-based SDE \cite{song2020score} proposed that the \textit{diffusion process} could be modeled as the solution to an Itô SDE as Eq. \ref{forward-SDE}

\begin{equation}
\label{forward-SDE}
\mathrm{d}\mathbf{x} = \mathbf{f}(\mathbf{x}, t)\mathrm{d}t + g(t)\mathrm{d}\mathbf{w}
\end{equation}
where the $\mathbf{w}$ is the standard Wiener process when time $t$ evolves from $0$ to $T$, $\mathbf{f}(\cdot, t): \mathbb{R}^d \to \mathbb{R}^d$ is a vector-valued function called the drift coefficient of $\mathbf{x}_t$, and $g(\cdot):\mathbb{R}\to \mathbb{R}$ is a scalar function called the diffusion coefficient of $\mathbf{x}_t$.

And the \textit{reverse process} is the solution of Eq. \ref{reverse-SDE}

\begin{equation}
\label{reverse-SDE}
\mathrm{d}\mathbf{x}=\left[  \mathbf{f}(\mathbf{x}, t)-g^2(t)\nabla_\mathbf{x} \log q_{t}(\mathbf{x}) \right] \mathrm{d}t +g(t)\mathrm{d}\bar{\mathbf{w}}
\end{equation}
where $\bar{\mathbf{w}}$ is a standard Wiener process when time $t$ flows backward from $T$ to $0$, and $\mathrm{d}t$ is an infinitesimal negative time step. With the notation that $q_t(\mathbf{x})$ is the probability density of $\mathbf{x}_t$,  the training objective is 

\begin{equation}
\arg \min_{\theta} \mathbb{E}_{t\in \mathcal{U}(0,T)} \mathbb{E}_{q_t(\mathbf{x})}
\left[
g^2(t)||\nabla_\mathbf{x} \log q_t(\mathbf{x})-s_{\theta}(\mathbf{x})||^2_2
\right]
\end{equation}

Although, DDPM and SMLD represent two different ways of adding noise, both can be represented by SDEs. For DDPM, the corresponding SDE is Eq. \ref{VP-SDE}, named the Variance Preserving SDE (VP-SDE) since it gives a process with bounded variance when $t\to \infty$. And for SMLD, the corresponding SDE is Eq. \ref{VE-SDE}, named the Variance Exploding SDE (VE-SDE) since it yields a process with exploding variance when $t\to \infty$.

\begin{equation}
\label{VP-SDE}
\mathrm{d} \mathbf{x} = - \frac{1}{2} \beta(t)\mathbf{x} \mathrm{d} t + \sqrt{\beta(t)} \mathrm{d}\mathbf{w} 
\end{equation}

\begin{equation}
\label{VE-SDE}
\mathrm{d} \mathbf{x} = \sqrt{\frac{\mathrm{d} \sigma^2(t)}{\mathrm{d}t}} \mathrm{d}\mathbf{w}
\end{equation}

And the relationship between SDE and ODE also represents the relationship between probabilistic and deterministic sampling. For a SDE in Eq. \ref{forward-SDE}, it can be shown to be equivalent to the following Eq. \ref{Fokker-Planck-equation}. Since $\sigma(t)\mathrm{d}\mathbf{w}$ can bring randomness if $\sigma(t) \ne 0$, this case denotes probabilistic sampling. And if the variance $\sigma(t) =0$, sampling contains no randomness and is therefore a deterministic process, then we can obtain an ODE as in Eq. \ref{reverse-ODE}. This deterministic sampling could also be viewed as a normalizing flow and used to estimate probability densities and likelihoods.

\begin{equation}
\label{Fokker-Planck-equation}
\mathrm{d} \mathbf{x}=\left[\mathbf{f}(\mathbf{x}, t)-\frac{g^2(t)-\sigma^2(t)}{2}\nabla_{\mathbf{x}} \log q_t(\mathbf{x})\right] \mathrm{d} \mathbf{t}+\sigma(t) \mathrm{d} \mathbf{w}
\end{equation}

\begin{equation}
\label{reverse-ODE}
\mathrm{d}\mathbf{x}=\left[  \mathbf{f}(\mathbf{x}, t)-\frac{1}{2}g^2(t)\nabla_\mathbf{x} \log q_{t}(\mathbf{x}) \right] \mathrm{d}t
\end{equation}

In addition, the author in \cite{song2020score} introduced new mathematical tools that offer the possibility of using DPMs to solve inverse problems in medical imaging \cite{song2021solving}.

\subsection{Relationship}
The emergence of Score-based SDE \cite{song2020score} provided us with a mathematical tool for the theoretical study of DPMs, which revealed useful relationships between discrete and continuous-time diffusion probabilistic models.

\paragraph{Noise Estimation and Score Matching}

 Score-based SDE \cite{song2020score} revealed a special relationship between the noise estimation $ \epsilon_\theta(\mathbf{x}_t, t)\approx \epsilon_t$ in DDPM and the score matching $s_\theta{(\mathbf{x}_t)} \approx \nabla_{\mathbf{x}_t} \log q(\mathbf{x}_t)$ in SMLD. Specifically, 
Eq. \ref{DDPM-forward-trans} indicates that $q(\mathbf{x}_t| \mathbf{x}_0) \sim \mathcal{N}(\sqrt{\bar{\alpha}_t}\mathbf{x}_0, (1-\bar{\alpha}_t)\mathbf{I})$, so we can calculate the gradient of log-likelihood $\nabla_{\mathbf{x}_t} \log q(\mathbf{x}_t| \mathbf{x}_0)$: 

\begin{align}
    \begin{aligned}
\label{noise-cond-score}
\nabla_{\mathbf{x}_t} \log q(\mathbf{x}_t| \mathbf{x}_0) = \nabla_{\mathbf{x}_t}
\left[
-\frac{(\mathbf{x}_t-\sqrt{\bar{\alpha}_t})^2}{2(1-\bar{\alpha}_t)\mathbf{I}}
\right]
 =\frac{-\sqrt{1-\bar{\alpha}_t} \epsilon_t}{1-\bar{\alpha}_t} = -\frac{\epsilon_t}{\sqrt{1-\bar{\alpha}_t}}\approx-\frac{\epsilon_\theta(\mathbf{x}_t, t)}{\sqrt{1-\bar{\alpha}_t}}
    \end{aligned}
\end{align}

Therefore, by considering the expectation of $\nabla_{\mathbf{x}_t} \log q(\mathbf{x}_t| \mathbf{x}_0)$ with respect to $q(\mathbf{x}_0)$ in Eq. \ref{noise-score-eq}. The score matching in Eq. \ref{score-matching} is equivalent to the noise estimation in Eq. \ref{DDPM-obj-simple} divided by a constant.

\begin{align}
\label{noise-score-eq}
\begin{aligned}
s_\theta{(\mathbf{x}_t)} \approx \nabla_{\mathbf{x}_t} \log q(\mathbf{x}_t)
=  \mathbb{E}_{q(\mathbf{x}_0)} [\nabla_{\mathbf{x}_t} q(\mathbf{x}_t \vert \mathbf{x}_0)]
 \approx \mathbb{E}_{q(\mathbf{x}_0)} \left[ - \frac{\epsilon_\theta(\mathbf{x}_t, t)}{\sqrt{1 - \bar{\alpha}_t}} \right]
= - \frac{\epsilon_\theta(\mathbf{x}_t, t)}{\sqrt{1 - \bar{\alpha}_t}}
\end{aligned}
\end{align}

\paragraph{Reverse Sampling and SDE(ODE) Solver}

Discrete-time DPMs can be formally viewed as discrete approximations of continuous-time SDEs, and sample generation of different DPMs corresponds to different differential equation solvers. Specifically, for the DDPM, Song et al. \cite{song2020score} stated that the sampling of its reverse process corresponds to the maximum likelihood SDE solver of the diffusion SDE, and Bao et al. \cite{bao2022analytic} gave an analytic form for the optimal variance of the process. For the DDIM, Song et al. \cite{song2020denoising} first illustrated the similarity between its iterative sampling and solving ODEs. Salimans and Ho \cite{salimans2022progressive} pointed out that sampling corresponds to the first-order ODE solver of the diffusion ODE after a certain transformation. Then, Lu et al. \cite{lu2022dpm} proved that DDIM is the first-order ODE solver based on diffusion ODEs with semilinear structure, and they also gave analytic solutions of the corresponding higher-order solvers.

\subsection{Conditional DPMs}
\label{conditional-generation}
DDIM \cite{song2020denoising} provided a way of conditional generation through deterministic sampling of noisy hidden variables, and the score-based SDE \cite{song2020score} pointed out that conditional generation can be achieved by solving a conditional reverse-time SDE and provided three examples of controllable generation, which opened up the study of conditional generation. The guided-DPMs \cite{dhariwal2021diffusion} then proposed training a noisy image classifier $q(y|\mathbf{x}_t)$ to control the generation of samples conditioned on the category $y$, using the gradient $\nabla_{\mathbf{x}_t}\log q(y|\mathbf{x}_t)$ with intensity $\gamma$. In contrast, the author in \cite{ho2022classifier} highlighted that category guidance can be achieved by introducing the condition $y$ during the training of diffusion probabilistic models, which is an implicit way of constructing a classifier that could adopt data pairs of conditional and perturbed images. Furthermore, recent works \cite{nichol2021glide, bansal2023universal,liu2023more} extended category conditions to encompass image, text, and multi-modal conditional generations. As another representative approach for conditional generation, the Latent Diffusion Probabilistic Models (LDMs) \cite{rombach2022high} considered constructing a pre-trained Encoder-Decoder and used DPMs to generate the hidden variables at the bottleneck, which reduced the computational complexity of DPMs and made it possible for conditional operations in the latent space.
\section{Emerging Applications in MRI}\label{Sec:Emerging-Application-in-MRI}
This section will focus on introducing the application of diffusion probabilistic models in Reconstruction (Section \ref{Reconstruction}), Image Generation (Section \ref{Image-Generation}), Image Translation (Section \ref{Multi-modal-Translation}), Segmentation (Section \ref{Segmentation}), Anomaly Detection (Section \ref{Anomaly-Detection}) and some other emerging applications (Section \ref{Other-Applications}).

\subsection{Reconstruction}\label{Reconstruction}
MRI acceleration which involves reconstructing undersampling data to remove artifacts is a popular research topic. Data-driven deep learning methods have achieved great success in MRI reconstruction, most of which are based on convolutional neural networks and require massive samples for training. DPMs have shown potential in solving the inverse problem of MRI reconstruction by obtaining better reconstruction quality and generalization capability. The studies adopting DPMs for MRI reconstruction are listed in Table \ref{tab:Reconstruction}, including the adopted DPM, the data domain where the DPM is applied, single- or multi-coil data, whether the fully-sampled data is required, and the code link.

\begin{table*}[htp]
\centering
\fontsize{9}{10}\selectfont{
\begin{threeparttable}[b]
\caption{Emerging DPMs in MRI Reconstruction.}

\label{tab:Reconstruction}

\begin{tabular}{cccccc}
\hline
 \multicolumn{1}{c}{Paper} & \multicolumn{1}{c}{Method} & \multicolumn{1}{c}{Domain} & \multicolumn{1}{c}{Coil}  & \multicolumn{1}{c} {FS Data\tnote{\#}} & \multicolumn{1}{c}{Code\tnote{*}} \\ \hline
\cite{korkmaz2023self} & DDPM & Image & Single/Multi & No & \href{https://github.com/yilmazkorkmaz1/ssdiffrecon}{\textit{link}} \\
  \cite{ravula2023optimizing} & SDE & Image & Multi & Yes & \href{https://github.com/sriram-ravula/mri_sampling_diffusion}{\textit{link}}\\
  \cite{aali2023solving} & SDE & Image & Multi & No & - \\
  \cite{cui2023spirit} & SDE & Image & Multi & Yes & - \\
  \cite{lee2023improving} & SDE & Image & Single & Yes & \href{https://github.com/hyn2028/tpdm}{\textit{link}}  \\
  \cite{chung2023fast} & DDIM & Image & Multi & Yes & - \\
  \cite{chung2023solving} & SDE & Image & Single & Yes & \href{https://github.com/HJ-harry/DiffusionMBIR}{\textit{link}} \\
  \cite{gungor2023adaptive} & DDPM & Image & Multi & Yes & \href{https://github.com/icon-lab/AdaDiff}{\textit{link}}  \\
  \cite{peng2023one} & SDE & K-space & Multi & Yes & \href{https://github.com/yqx7150/HKGM}{\textit{link}}  \\
  \cite{Luo_2023} & SDE & Image & Single/Multi & Yes & \href{https://github.com/mrirecon/spreco}{\textit{link}}  \\
  \cite{cao2022accelerating} & DDPM & K-space & Single & Yes & - \\
  \cite{cui2022self} & SDE & Image & Single & No & -\\
  \cite{peng2022towards} & DDPM & Image & Single & Yes & \href{https://github.com/cpeng93/DiffuseRecon}{\textit{link}} \\
  \cite{xie2022measurement} & DDPM & K-space & Single & Yes & \href{https://github.com/Theodore-PKU/MC-DDPM}{\textit{link}}  \\
  \cite{cao2022high} & SDE & Image & Multi & Yes & - \\
  \cite{tu2022wkgm} & SDE & K-space & Multi & Yes & \href{https://github.com/yqx7150/wkgm}{\textit{link}} \\
  \cite{chung2022score} & SDE & Image & Single/Multi & Yes & \href{https://github.com/HJ-harry/score-MRI}{\textit{link}}  \\
 \cite{chung2022come} & SDE & Image & Single &  Yes & -  \\
  \cite{song2021solving} & SDE & Image & Single & Yes & \href{https://github.com/yang-song/score_inverse_problems}{\textit{link}} \\
  \cite{jalal2021robust} & SDE & Image & Multi & Yes & \href{https://github.com/utcsilab/csgm-mri-langevin}{\textit{link}} \\
 \hline
\end{tabular}
\begin{tablenotes}
    \item[\#] "FS Data" means the fully-sampled data is needed for training. 
    \item[*]"-" indicates that the code is not available.
\end{tablenotes}
\end{threeparttable}
}
\end{table*}
The forward MR acquisition model can be formulated as:
\begin{equation}
    \mathbf{y}=\mathbf{Ax}+\boldsymbol{\epsilon}
    \label{MR forward model}
\end{equation}
where $\mathbf{y}$ is the acquired k-space data, $\mathbf{x}$ is the imaging object, $\mathbf{A=MFS}$ is the encoding operator, with $\mathbf{M}$ being the undersampling mask, $\mathbf{F}$ indicating the Fourier transform, $\mathbf{S}$ denoting the coil sensitivity maps and $\boldsymbol{\epsilon} \sim \mathcal{N}(0, \sigma^2_\epsilon\mathbf{I})$ is the acquisition noise. Reconstructing MR image from the undersampled k-space data $\mathbf{y}$ is commonly formulated as optimizing the following problem: 
\begin{equation}
    \mathbf{x}^*=\underset{\mathbf{x}}{\arg \min} \frac{1}{2}\|\mathbf{Ax}-\mathbf{y}\|_2^2+\mathcal{R}(\mathbf{x})
    \label{MR optimization problem}
\end{equation}
where the first term enforces data consistency and $\mathcal{R}(\mathbf{x})$ is the regularization term to stabilize the solution.

Based on the Score-based SDE framework, MR images can be sampled from the posterior distribution through the reverse-time SDE as Eq. \ref{conditional reverse SDE}, which is different from traditional methods that directly modeling the prior information of $\mathbf{x}$ as in the reconstruction optimization as Eq. \ref{MR optimization problem}.
\begin{equation}
\mathrm{d} \mathbf{x}=[\mathbf{f}(\mathbf{x}, t)-g(t)^2\nabla_{\mathbf{x}} \log p(\mathbf{x}_{t} |\mathbf{y})] \mathrm{d} t+g(t) \mathrm{d} \overline{\mathbf{w}}
\label{conditional reverse SDE}
\end{equation}
According to the Bayes' rule, we have that:
\begin{equation}
   \nabla_{\mathbf{x}} \log p(\mathbf{x}_{t} | \mathbf{y}) = \nabla_{\mathbf{x}} \log p(\mathbf{x}_{t})+ \nabla_{\mathbf{x}} \log p(\mathbf{y}| \mathbf{x}_{t})
\label{bayesian trans}
\end{equation}
The first term in Eq. \ref{bayesian trans} is the score function of the prior distribution, which can be estimated via score-matching. The second term is the likelihood, which has no closed-form solution as there is no explicit dependency of $\mathbf{y}$ on $\mathbf{x}_{t}$. 

There are different ways of approximating the likelihood term. Jalal et al. \cite{jalal2021robust} proposed to utilize an approximation $\nabla_{\mathbf{x}} \log p(\mathbf{y}|\mathbf{x}_{t}) \approx  \frac{\mathbf{A}^{H} (\mathbf{y} - \mathbf{Ax}_{t} ) }{\sigma_{\epsilon}^{2}} $, which is valid when $\boldsymbol{\epsilon}$ is Gaussian noise with variance of $\sigma_{\epsilon}^{2}$ and $t=0$. For higher noise perturbation levels $t \rightarrow T$, $\nabla_{\mathbf{x}} \log p(\mathbf{y}|\mathbf{x}_{t}) \approx  \frac{\mathbf{A}^{H} (\mathbf{y} - \mathbf{Ax}_{t} ) }{\sigma_{\epsilon}^{2}+ \lambda_{t}^{2}}$, where $\{ \lambda\}_{t=1}^{T}$ are hyperparameters \cite{chung2023diffusion}. 

\begin{figure}[htbp]
	\centering
    \includegraphics[width=1\linewidth]{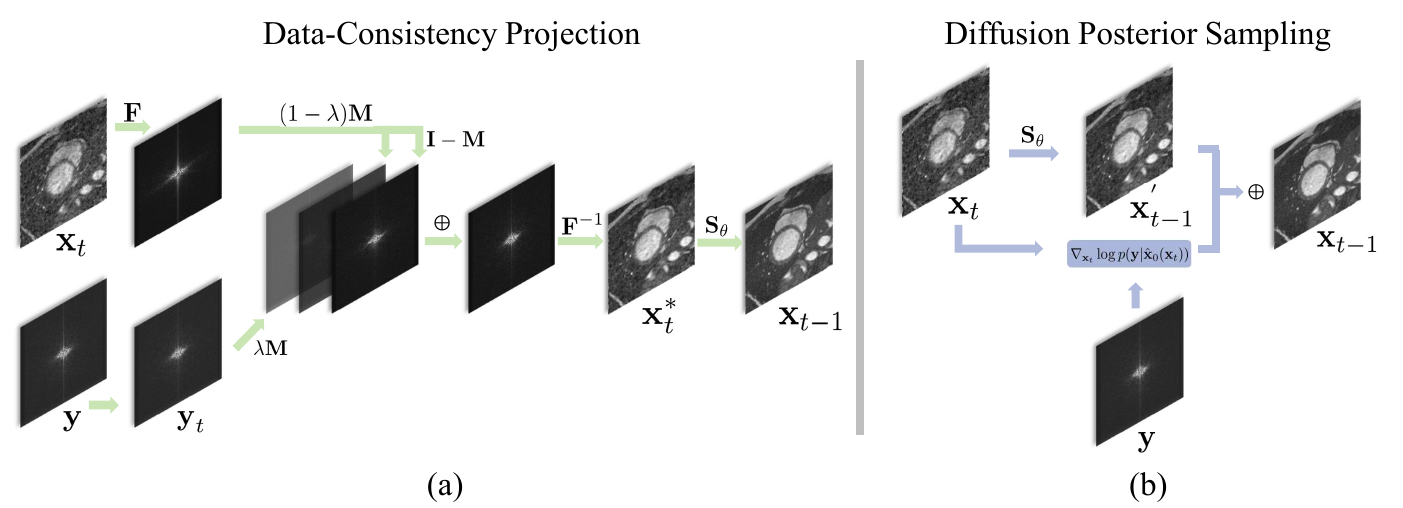}

	\caption{Two different approaches to realize conditional sampling during the reverse diffusion process of DPMs in MRI reconstruction. It shows how to sample $\mathbf{x}_{t-1}$ from $\mathbf{x}_{t}$ with the acquired data. }
     \label{fig:Sec3_Reconstruction}
\end{figure}

Moreover, Song et al.\cite{song2021solving} and Chung et al. \cite{chung2022score} proposed to perform unconditional sampling based on Eq. \ref{reverse-SDE} firstly, and then project the intermediate sampling result to the measurement space so that the data-consistency can be performed for the intermediate generation, $\mathbf{x}_{t}$:
\begin{equation}
\label{data-consistency enforcement}
    \mathbf{x}_{t}^{*}=\mathbf{F}^{-1}\left[\lambda \mathbf{My}_{t}+(1-\lambda)\mathbf{MF}\hat{\mathbf{x}}_{t}+(\mathbf{I-M})\mathbf{F}\hat{\mathbf{x}}_{t}\right]
\end{equation}
where $\mathbf{y}_{t}$ is the noise-corrupted acquired data obtained by disturbing $\mathbf{y}$ in the same way to that in the forward process, $\mathbf{\hat{x}_{t}}$ is the unconditional sampling result, and $\lambda$ is the hyperparameter balancing the data-consistency and the unconditional generation. Fig. \ref{fig:Sec3_Reconstruction}(a) illustrates the data-consistency enforcement in Eq. \ref{data-consistency enforcement}.

Furthermore, Chung et al. \cite{chung2023diffusion} proposed the Diffusion Posterior Sampling (DPS) to approximate $\nabla_{\mathbf{x}} \log p(\mathbf{x}_{t}|\mathbf{y})$  by exploiting the result from Tweedie's rule. For the case of VE-DPMs, $p_{0 t}(\mathbf{x}_t |\mathbf{x}_0)=\mathcal{N}(\mathbf{x}_t ; \mathbf{x}_0,\sigma^2_{t}\mathbf{I})$, we can obtain the closed-form expression for the expectation of posterior: 
\begin{equation}
\label{eq:tweedie}
\begin{aligned}
    \mathbf{\hat{x}}_{0} = \mathbb{E}_{ \mathbf{x}_t \sim p_t(\mathbf{x}_t|\mathbf{x}_0)}\left[\mathbf{x}_0 | \mathbf{x}_t\right]
     = \mathbf{x}_t + \sigma_t^2 \nabla_{\mathbf{x}_t} \log p_t(\mathbf{x}_t) 
    \approx \mathbf{x}_t + \sigma_t^2 s_{\theta}(\mathbf{x}_{t}, t)
\end{aligned}
\end{equation}
Eq. \ref{eq:tweedie} means the expectation of posterior can be approximated by the trained score-based model $s_{\theta}(\mathbf{x}_{t}, t)$. Hence, the likelihood term can be approximated by $\nabla_{\mathbf{x}} \log p(\mathbf{y}|\mathbf{x}_{t}) \approx \nabla_{\mathbf{x}} \log p(\mathbf{y} | \mathbf{\hat{x}}_{0}(\mathbf{x}_{t}))$. With the DPS approximation in Fig. \ref{fig:Sec3_Reconstruction}(b), Eq. \ref{conditional reverse SDE} can be used to reconstruct MR images. 

The DPM-based MRI reconstruction methods can be generally categorized according to the domain (image or k-space) they are applied to, which will be introduced separately in the following.

\paragraph{DPMs in image domain}
Most DPMs are applied in the image domain for MRI reconstruction. Jalal et al. \cite{jalal2021robust} first proposed training a score-based model on MR images as a prior for MRI reconstruction, which generated high-quality images through Langevin dynamics posterior sampling and showed superior performance in comparison with the end-to-end supervised learning method. 
Furthermore, Luo et al. \cite{Luo_2023} provided a more detailed analysis of the robustness and flexibility of DPMs for reconstructing MR images, elucidating the reconstruction uncertainty and the computational burden.

To achieve conditional generation, Chung et al. \cite{chung2022score} proposed a conditional sampling method given measurements, which added a consistency mapping between the predictor and corrector during the sampling process. Their method can also be applied to multi-coil k-space data by reconstructing each coil image separately, followed by a sum-of-squares coil combination. Song et al.\cite{song2021solving} trained a score-based model on fully-sampled images to capture the prior distribution, and they provided detailed mathematical descriptions of how to incorporate acquired measurements and the known physics model into an unconditional sampling process. The basic idea is to project the unconditionally sampled images at each diffusion time step to make them consistent with $\mathbf{y}_{t}$ as in Eq. \ref{data-consistency enforcement}. Peng et al. \cite{peng2022towards} followed the idea of adding data-consistency projection in the sampling phase while shortening the reconstruction schedule, and averaged multiple reconstructions at each diffusion time step to avoid the degradation of reconstruction quality caused by shortening the sampling schedule. Güngör et al. \cite{gungor2023adaptive} proposed AdaDiff, which adopted a large step size to accelerate the sampling process and generate the initial reconstruction, which was refined in the adaption phase by comparing with the reference data.

Differing from the above works utilizing fully-sampled images in the forward diffusion process, there are recent studies demonstrating the feasibility of training MRI reconstruction DPMs with only undersampled MR images. Cui et al. \cite{cui2022self} utilized a Bayesian neural network to learn the prior data distribution from undersampled images, and then perturbed the distribution and trained a score-based model to reconstruct MR images. Aali et al. \cite{aali2023solving} proposed a novel loss function to train the score-based model by combining Stein's unbiased risk estimate with denoising score matching. This method was able to jointly denoise noisy data disturbed by Gaussian noise and train the score-based model. Korkmaz et al. \cite{korkmaz2023self} employed a k-space masking strategy for self-supervised learning of DPMs, where the undersampled k-space data was randomly divided into two parts which were respectively used for data consistency and calculating the reconstruction loss. Furthermore, an unrolled transformer network was designed in this work to replace the commonly used denoising U-Net, which consists of a mapper network and an unrolled denoising block. The mapper network was used to capture encoding information of time and prior information extracted from under-sampled images. Denoising blocks were used for image denoising and performing data consistency.

Besides learning directly from undersampled data, recent developments of DPMs for MRI reconstruction also focus on improving the forward and reverse processes of SDE. Cao et al. \cite{cao2022high} proposed HFS-SDE to achieve more stable and faster MRI reconstruction by restricting the diffusion process to the high-frequency region.  Cao et al. \cite{cao2022spiritdiffusion} and Cui et al. \cite{cui2023spirit} proposed a new paradigm for the SDE design for multi-coil reconstruction by replacing the drift coefficient in the original SDE with the gradient of the self-consistent term in SPIRiT \cite{lustig_spirit_2010}, a parallel imaging MRI reconstruction method, and enforced the self-consistent property of the Gaussian noise of the diffusion coefficient.

Instead of using the reverse SDE to reconstruct MR images, Ravula et al. \cite{ravula2023optimizing} attempted to optimize the undersampling pattern described by a Bernoulli distribution with learnable parameters. The optimization is conducted via minimizing the error between fully sampled signal and the result generated by a score estimator conditioned on the corresponding undersampled signal. In addition, DPMs have been specifically designed for 3D MRI reconstruction. Chung et al. \cite{chung2023solving} proposed DiffusionMBIR, where a 2D DPM was used to perform the reconstruction slice-by-slice, and then the classical Total Variance prior was added along the slice direction, which enhance the intrinsic coherence between the slice-wise reconstructions. Furthermore, Lee et al. \cite{lee2023improving} proposed to utilize two perpendicular pre-trained 2D DPMs to enhance the exploiting of 3D prior distribution.

\paragraph{DPMs in k-space}
Among the MRI reconstruction DPMs applied in the k-space domain, the most representative ones are MC-DDPM \cite{xie2022measurement} and CDPM \cite{cao2022accelerating}. MC-DDPM defined the diffusion process in k-space and added the under-sampling mask to the conditional distribution. Therefore, this method uses measurement priors to ensure data consistency in the sampling process. In addition, this method could provide an assessment of the uncertainty in the sampling results. CDPM leveraged the undersampling mask and the observed k-space data as the conditions of the forward Markov chain, based on which learned the distribution of the k-space data that was not acquired. Tu et al. \cite{tu2022wkgm} proposed WKGM to achieve the multi-coil reconstruction in k-space by weighting the initial k-space data to lift high-frequency and suppress low-frequency data. The advantage of this weighting is to reduce the dynamic range of the k-space magnitude, so the prior distribution can be well captured.

\subsection{Image Generation}\label{Image-Generation}

Because DPMs can steadily generate MRI images with specific data structures and pathological features, DPMs also provide a new approach for data augmentation, which can help address the limitations of the scarcity of MRI datasets in downstream diagnostic models. Specifically, how to utilize DPMs to handle complex data formats including but not limited to 2D, 3D, and spatiotemporal data, and how to identify and apply meaningful prior to obtain samples that meet practical requirements, are the major issues that need to be addressed by DPMs for image generation in MRI. Table \ref{tab:Image-Generation} summarizes the related works, including the adopted DPM, the target organ, the data dimension and generation conditions associated with the method, the specific task of the method and its potential application discussed in the article, and the code link.

\begin{table*}[htp]
\centering
\fontsize{9}{10}\selectfont{
\begin{threeparttable}
\caption{Emerging DPMs in MRI Generation.}
\label{tab:Image-Generation}

\begin{tabular}{ccccccc}
\hline
\multicolumn{1}{c}{Paper} & Method &\multicolumn{1}{c}{Organ} & \multicolumn{1}{c}{Dimension} & Condition & Specific Task | Application&\multicolumn{1}{c}{Code\tnote{*}} \\ 
\hline
 \cite{pan20232d} &  DDPM & Cardiac  & 2D & Unconditional & Structural Design | Classification& - \\
  \cite{fernandez2022can} & LDM &Brain & 2D& Label Generator & Image-label Generation | Segmentation& - \\
  \cite{de2023medical} & LDM &Prostate & 2D& Textual Inversion &Appearance Control | Classification& - \\
  \cite{dorjsembe2022three} & DDPM & Brain & 3D&Unconditional & Structural Design | 3D Generation& \href{https://github.com/DL-Circle/3D-DDPM}{\textit{link}} \\
  \cite{dorjsembe2023conditional} & DDPM & Brain &3D& Mask Prior & Structural Design | Segmentation& \href{https://github.com/mobaidoctor/med-ddpm}{\textit{link}} \\
  \cite{han2023medgen3d} & DDPM &Brain & 3D & Mask Prior & Image-label Generation | Segmentation& - \\
  \cite{durrer2023diffusion} & DDPM & Brain &3D & Anatomical Prior & Contrast Harmonization | Segmentation& - \\
  \cite{peng2022generating} & DDPM &Brain  & 3D &Slice Prior & Structural Design | 3D Generation& - \\
  \cite{pinaya2022brain} & LDM &Brain & 3D & Covariates Prior & Structural Design | Data Resource& -  \\
  \cite{khader2022medical} & LDM & \makecell[c]{Brain\\Brest\\Knee} & 3D & Unconditional & Structural Design | Segmentation & \href{https://github.com/FirasGit/medicaldiffusion}{\textit{link}} \\
  \cite{kim2022diffusion} & DDPM & Cardiac &4D &  Deformation & Deformation Estimation | 4D Generation& \href{https://github.com/torchddm/ddm}{\textit{link}} \\
  \cite{yoon2023sadm} & DDPM & \makecell[c]{Cardiac\\Brain} & 4D & Classifier-Free Guidance& Structural Design | 4D Generation& \href{https://github.com/ubc-tea/sadm-longitudinal-medical-image-generation}{\textit{link}} \\
  \cite{akbar2023beware} & DDPM  &Brain & 2D & Unconditional & Data Memorization | Feasibility& - \\
  \cite{dar2023investigating} & LDM & Knee & 3D &Unconditional& Data Memorization | Feasibility& - \\
\hline
\end{tabular}
\begin{tablenotes}
\item[*]"-" indicates the code is not available.\\
\end{tablenotes}
\end{threeparttable}
}
\end{table*}

Due to the computationally expensive nature of early DPMs, their application in MR image generation initially starts from 2D MRI. Such works primarily focused on optimizing the structure of noise estimators, incorporating guidance mechanisms of generation, and exploring their potential applications in downstream tasks. Pan et al. \cite{pan20232d} proposed using the Swin-vision transformer \cite{liu2021swin} as the noise estimator for DDPM to capture local and global information of noisy hidden variables, which include abundant details of the experimental setup, and further discussed the effect of the generated data in classification tasks. The development of the Latent Diffusion Models(LDMs) \cite{rombach2022high}, brings new vitality to the generation of 2D MRI data. To improve the training of the segmentation models, Fernandez et al. \cite{fernandez2022can} proposed the brainSPADE consisting of a synthetic label generator in spatial latent space using LDM and an encoder-decoder-based semantic image generator. Moreover, the study in \cite{de2023medical} discussed the feasibility of fine-tuning a LDM trained on natural images for medical imaging applications. The innovation of this work lies in using textual inversion to control the intensity of variables in the latent space of the LDM. Combining with the hidden states that represent different diseases, it was able to generate diverse samples with various types of diseases and severity levels, and demonstrated the potential to control the appearance of lesions by manipulating the segmentation masks.

Before the application of LDMs in 3D MRI generation, DPMs were often used to solve the 3D generation problem by first generating 2D sub-slices and then assembling them into a 3D volume. Dorjsembe et al. \cite{dorjsembe2022three} first reported the adoption of a 3D DDPM for generating 3D Brain MR images by replacing all 2D operations into 3D ones. Dorjsembe et al. \cite{dorjsembe2023conditional} then introduced a method based on 3D DDPM  in synthesizing volumes conditioned on a given segmentation mask, which also demonstrated the effectiveness in enhancing the performance of segmentation models. However, they still face the challenge of computationally expensive 3D operations. Therefore, Han et al. \cite{han2023medgen3d} represented 3D volumetric data as 2D sequences to use MC-DPM to generate mask sequences that conform to the anatomical geometry, and then designed a conditional generator to synthesize 3D MRI images corresponding to the mask sequence. Durrer
et al. \cite{durrer2023diffusion} applied DDPM on a paired 3D MRI dataset with scanner-inherent differences in a 2D subslice way, which generated images that retain anatomical information but have adjusted contrasts, thus increasing the comparability between scans with different contrasts by mapping images into the same target contrast. To enforce the inter-slice dependency of generated 3D brain MRIs,  Peng et al. \cite{peng2022generating} designed a strategy to calculate the attention weights for MRI volume generation using slice-wise masks in the DPM.

Although the 3D MRI generation approaches with 2D sub-slice operations reduced the spatial complexity, they prolonged the generation time. Furthermore, these methods may suffer from producing generation artifacts and contrast variations when trained inappropriately with small data samples. LDMs provided an alternative approach for the controllable generation of 3D MRI. Pinaya et al. \cite{pinaya2022brain} used LDMs to create 3D synthetic MRI images of the adult brain and leveraged the cross-attention mechanism to incorporate covariates (e.g., age, gender, and brain volume) to make the generations conform to expected representations. Following the idea of LDMs, Khader et al. \cite{khader2022medical} used a pre-trained VQ-GAN \cite{esser2021taming} to encode images into a low-dimensional latent space and constructed a DDPM on the latent representations to generate 3D samples. The generation results are also applied to train a segmentation network. 

Applying DPMs to the generation of high-dimensional MRI such as dynamic MRI remains a pressing frontier issue that needs to be addressed. There are some pioneering studies in this regard. Kim et al.\cite{kim2022diffusion} combined DPMs with traditional deformable deep learning models to generate the intermediate frames of Cardiac Cine MRI. Moreover, to add time dependence to the generation of DPMs for multi-frame cardiac MRI and longitudinal brain MRI, Yoon et al. \cite{yoon2023sadm} introduced a sequence-aware transformer that can combine the time information into the classifier-free guidance training, which facilitate the generation of missing frames and future images in longitudinal studies.

Furthermore, a critical challenge in DPMs for MRI generation lies in whether the samples generated by DPMs benefit downstream tasks. Akbar et al. \cite{akbar2023beware} argued that commonly used metrics such as Fréchet inception distance and Inception Score are not sufficient to judge whether the generated results of DPMs duplicate the training data. Therefore, this study explored the synthesis ability of DPMs on brain tumor MR images and concluded that DPMs were more likely to memorize the training images than GANs, especially with small-size training datasets. As a further study, Dar et al.\cite{dar2023investigating} constructed a self-supervised model based on the contrastive learning approach that compared the generation and training samples on their low-dimensional latent representations. This research achieved a similar conclusion that DPMs may memorize the training data.

\subsection{Image Translation}\label{Multi-modal-Translation}

Image translation, as a useful way of exploring the relationship between medical image modalities such as MRI, CT and PET, can enrich the available imaging modalities for downstream medical image analysis tasks. However, the establishment of generative models to achieve medical image translation remains a challenge due to the high cost of acquiring different modality images and the complex nonlinear relationships between signals of different modalities. Recently, thanks to the advancements both in principles and methods for cross-domain representation in DPMs \cite{su2022dual,meng2021sdedit,zhao2022egsde}, the use of DPMs for MRI image translation is attracting increasing attention. Table \ref{tab:Image-Translation} summarizes the related works of DPMs in MRI translation, including the adopted DPM, the target organ, the source and target modalities, the type of the translation task, whether model training requires paired data, and the code link.

\begin{table*}[htp]
\centering
\fontsize{9}{10}\selectfont{
\begin{threeparttable}
\caption{Emerging DPMs in MR image translation.}
\label{tab:Image-Translation}
\begin{tabular}{ccclccc}
\hline
\multicolumn{1}{c}{Paper} & Method & \multicolumn{1}{c}{Organ} & \multicolumn{1}{c}{Source $\to$ Target} & Translation & Paired Training Data& \multicolumn{1}{c}{Code\tnote{*}} \\ \hline
 \cite{saeed2023bi} & DDPM & Prostate &  T2 $\to$ DW & 1-to-1 & Yes & - \\
 \cite{li2023ddmm} & DDIM & \makecell[c]{Pelvic\\Brain}  & MRI(T2) $\to$ CT & 1-to-1 & Yes& - \\
  \cite{taofeng2022brain} & SDE & Brain & MRI(T1) $\to$ PET & 1-to-1 & Yes& - \\
  \cite{zhao2023ddfm} & DDPM & Brain & \makecell[l]{MRI(T2) $\to$ CT, PET\\MRI(T2) $\to$ SPECT}& 1-to-1 & Yes & \href{https://github.com/Zhaozixiang1228/MMIF-DDFM}{\textit{link}} \\
  \cite{ozbey2023unsupervised} & DDPM & \makecell[c]{Pelvic\\Brain} & \makecell[l]{T1 $\rightleftharpoons$ PD\\T1 $\rightleftharpoons$ FLAIR\\MRI(T2) $\to$ CT} & 1-to-1 & No & \href{https://github.com/icon-lab/syndiff}{\textit{link}} \\
  \cite{wang2023zero} & SDE & \makecell[c]{Prostate\\Brain} & \makecell[l]{T1 $\rightleftharpoons$ T2\\T1 $\rightleftharpoons$ PD\\T2 $\rightleftharpoons$ PD\\T1 $\rightleftharpoons$ FLAIR\\T2 $\rightleftharpoons$ FLAIR\\ MRI(T1, T2) $\to$ CT} & 1-to-1 & No& - \\
  \cite{pan2023cycle} & DDPM & Brain & \makecell[l]{T1 $\rightleftharpoons$ T2\\T1 $\rightleftharpoons$ FLAIR} &  1-to-1 & Yes& - \\
  \cite{meng2022novel} & SDE & Brain & \makecell[l]{$C^1_4(\text{T1,T1ce,T2,FLAIR})$\tnote{\#}} & M-to-1\tnote{\&} & Yes& - \\ 
  \cite{jiang2023cola} & LDM & Brain & \makecell[l]{$C^1_4(\text{T1,T1ce,T2,FLAIR})$\tnote{\#}\\$C^1_3(\text{T1,T1ce,PD})$\tnote{\#}}& M-to-1\tnote{\&} & Yes& - \\
\hline
\end{tabular}
\begin{tablenotes}
    \item[*]"-" indicates the code is not available.\\
    \item[\#]"$C^1_n$" means this method selects one of the set of $n$ modalities as the target modality and the remaining $n-1$ modalities as the source modality. For example, $C^1_4(\text{{T1,T1ce,T2,FLAIR}})$ represents four different translation settings: T1,T1ce,T2 $\to$ FLAIR, FLAIR,T1ce,T2 $\to$ T1, FLAIR,T1,T2 $\to$ T1ce, and FLAIR,T1,T1ce $\to$ T2.\\
    \item[\&]"M-to-1" refers to the multi-to-one image translation.
\end{tablenotes}
\end{threeparttable}
}
\end{table*}
\begin{figure}[h]
	\centering
    \includegraphics[width=0.8\linewidth]{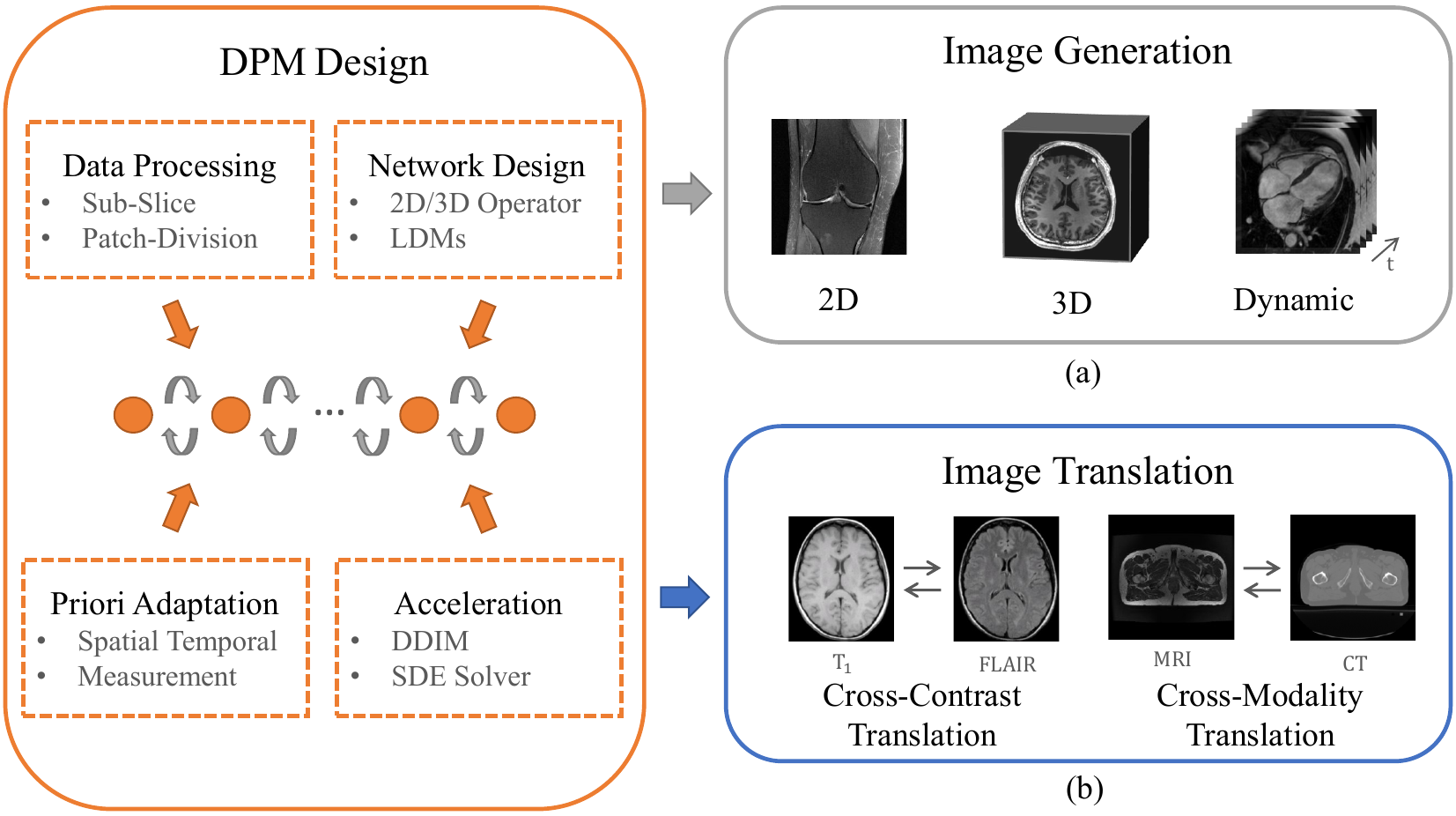}	
 \caption{Model design and common tasks of DPMs in MR image generation (a) and image translation (b).}
 \label{fig:Sec3_GeneTransProcess}
\end{figure}

Different from the image generation tasks which can be conditional and unconditional, as shown in Fig. \ref{fig:Sec3_GeneTransProcess}, image translation tasks mostly perform conditional generation, aiming to learn the underlying correlations between the source and target modality data distributions so that the missing modality image can be generated given the source modality. 

In other words, image generation explores the conditional relation within one distribution, while translation focuses on discovering the correlation between different distributions.

Starting with the one-to-one image translation, the early attempt to apply DPMs in MR image translation was to utilize data from a single source modality as a condition in the sampling process to generate the target modality. One way is to use an encoder to obtain the latent representation of the source modality, which is then combined with the noise estimator to achieve conditional generation. Saeed et al. \cite{saeed2023bi} encoded T2-weighted images through the BERT tokenizer as a condition acting on the middle layers of the noise estimator in synthesizing prostate diffusion-weighted MR images. Besides the latent representation, the original image can also be used as the generation condition. To synthesize high-fidelity CT images from MR images, Li et al. \cite{li2023ddmm} combined the DDIM guided by the MR image with a regularization term of the Range-Null Space Decomposed CT measurements in the sampling process.

The one-to-one image translation DPMs have been optimized in recent studies. Taofeng et al. \cite{taofeng2022brain} proposed to use the joint probability distribution of diffusion model (JPDDM) to synthesize brain PET images using ultrahigh field MRI (e.g., 5T MRI and 7T MRI) as the guidance. Zhao et al. \cite{zhao2023ddfm} redesigned the posterior sampling of DDPM with an unconditional generation and a conditional likelihood correction using the EM algorithm in natural image translation and then applied this method in generating CT images from MRI. Moreover, in order to achieve unsupervised training in the unpaired datasets, Özbey et al. \cite{ozbey2023unsupervised} proposed the SynDiff to incorporate adversarial modules within the DPM to form a cycle-consistent architecture, which first generated initial translations using a non-diffusion module containing two generator-discriminator pairs and then used the initial estimations as conditions for the diffusion module in generation. As an improvement, Wang et al. \cite{wang2023zero} proposed an unsupervised learning method named MIDiffusion to leverage a score-based SDE with an embedded conditioner , which can exploit local mutual information between target and source images to capture the identical cross-modality features without direct mapping between domains.

Despite the outstanding achievements of DPMs in 2D MRI translation by introducing different modality conditions and optimizing model architectures in DPMs, applying them to high-dimensional MRI data has yet to be extensively studied. Pan et al. \cite{pan2023cycle} developed a cycle-guided DDPM that used two 3D DDPMs to represent two different MRI contrasts. Exchanging the noisy latent variables in each timestep served as a latent code regularization to match the two MRI modalities in generation. Although this method reduced the uncertainty of the sampling process, how to design a more efficient DPM for 3D MRI translation remains an open question.

Compared with the one-to-one image translation, the many-to-one/many image translation tasks are more complex and require a particular model design. Meng et al. \cite{meng2022novel} developed the multi-modal completion framework in which a unified multi-modal conditional score-based generative model was proposed to generate the missing modalities using a multi-input multi-output conditional score network to learn the multi-modal conditional score of the multi-modal distributions. Also, Jiang et al. \cite{jiang2023cola} proposed a conditional LDM-based many-to-one generation model for multi-contrast MRI. This method used a similarity cooperative filtering mechanism to avoid over-compressing information in the latent space. The structural guidance and auto-weight adaptation strategies were adopted to synthesize high-quality images. Developments of more efficient operations in the latent space and the domain translation-related DPM structures could contribute to improving the DPMs performance in complex MRI translation tasks.


\subsection{Segmentation}\label{Segmentation}
Image segmentation, which aims to divide an image into different regions of interest, is a crucial step in medical image analysis applications. Manual segmentation is still considered as clinical standard, while it is noted that annotations made by multiple experts can vary significantly due to differences in experience, expertise, and subjective judgments. Deep learning methods have achieved state-of-the-art performance in medical image segmentation tasks. Recent studies have also found DPMs hold potential in this discriminative task because of their strong performance. Table \ref{tab:Segmentation} summarized the related segmentation DPMs in MRI, where the adopted DPMs, the target organ, the learning strategy, the different type of the segmentation masks, the condition associated with the method, and the code link are included. Specifically, we take whether the training of DPMs directly exploits each sample's annotation as the learning strategy criteria.

\begin{table*}[h]
\centering
\fontsize{9}{10}\selectfont{
\begin{threeparttable}
    \caption{Emerging DPMs in MRI Segmentation.}
    \label{tab:Segmentation}
\begin{tabular}{cccccccc}
\hline 
Paper & Method & Organ & Dimension &  Strategy & Segmentation & Condition & Code\tnote{*} \\ \hline
  \cite{amit2023annotator} & DDPM & \makecell[c]{Brain\\Prostate} & 2D & Supervised & Binary & Image & \href{https://github.com/tomeramit/Annotator-Consensus-Prediction}{\textit{link}} \\
  \cite{hu2023conditional} & DDIM & \makecell[c]{Brain\\Kidney} &2D &Weekly-supervised&  Binary &Classification Label & \href{https://github.com/xhu248/cond_ddpm_wsss}{\textit{link}}   \\
  \cite{akbar2023brain} & DDPM & Brain & 2D & Supervised & Binary & Mask & \href{https://github.com/muhamadusman/Assist/}{\textit{link}} \\
  \cite{alshenoudy2023semi} & DDPM & Brain &2D & Semi-supervised &Binary& Pixel-Level Classifier & \href{https://github.com/risc-mi/braintumor-ddpm}{\textit{link}}  \\
  \cite{tursynbek2023unsupervised} & DDPM & Brain  & 2D & Unsupervised & Binary &Unconditional &- \\
  \cite{rahman2023ambiguous} & DDPM & Brain & 2D & Supervised & Binary &Image and Ambiguity & \href{https://github.com/aimansnigdha/Ambiguous-Medical-Image-Segmentation-using-Diffusion-Models}{\textit{link}}  \\
  \cite{BerDiff} & DDPM & Brain &2D & Supervised & Binary &Image & -  \\
  \cite{bieder2023diffusion} & DDPM & Brain &3D & Supervised & Binary &Volumetric Data &- \\
  \cite{Diff-UNet} & DDIM & Brain  & 3D & Supervised & Binary & Volumetric Data &\href{https://github.com/ge-xing/diff-unet}{\textit{link}} \\
  \cite{fu2023importance} & DDPM & Prostate & 3D & Supervised & Multiplex &Volumetric Data & \href{https://github.com/mathpluscode/imgx-diffseg}{\textit{link}}\\
  \cite{MedSegDiff-V2} & DDPM & Brain & 2D & Supervised & Multiplex & Anchor and Semantic Condition &\href{https://github.com/WuJunde/MedSegDiff}{\textit{link}} \\
  \cite{MedSegDiff} & DDPM & Brain & 2D & Supervised & Multiplex &FF-Parser &\href{https://github.com/WuJunde/MedSegDiff}{\textit{link}}\\
  \cite{2022Pre-segmentation} & DDPM & Brain &2D & Supervised &Binary & Image and Pre-segmentation & -\\
 \cite{wolleb2022diffusion} & DDPM & Brain & 2D & Supervised & Binary &  Image  & \href{https://github.com/juliawolleb/diffusion-based-segmentation}{\textit{link}}\\
\hline
\end{tabular}
\begin{tablenotes}
    \item[*]"-" indicates the code is not available.
\end{tablenotes}
\end{threeparttable}
}
\end{table*}

Inspired by the remarkable success of DPMs in generating semantically valuable pixel-wise representations, Wolleb et al.\cite{wolleb2022diffusion} first introduced DDPM for brain MR image segmentation. They provided a scheme for DPM-based image segmentation by synthesizing the labeled data and obviating the necessity for pixel-wise annotation. Although pioneering, this method is extremely time-consuming. Guo et al. \cite{2022Pre-segmentation} proposed PD-DDPM to accelerate the segmentation process by using pre-segmentation results and noise predictions based on forward diffusion rules. This method outperformed previous DDPM even with fewer reverse sampling steps when combined with Attention-Unet. MedSegDiff \cite{MedSegDiff} improved DPM for medical image segmentation by proposing a dynamic conditional encoding strategy, eliminating the negative effect of high-frequency noise components via an FF-Parser. Subsequently, to achieve a better convergence between noise and semantic features, they proposed MedSegDiff-V2 \cite{MedSegDiff-V2} in which a transformer-based architecture combined with a Gaussian spatial attention block was used for noise estimation.

BerDiff model \cite{BerDiff} distinguished itself by using Bernoulli noise as the diffusion kernel, improving the DPM's accuracy for binary image segmentations, especially for discrete segmentation tasks. It can also efficiently sample the sub-sequences from the reverse diffusion trajectory, thus fastening the segmentation process. Collectively intelligent medical diffusion model, proposed by \cite{rahman2023ambiguous}, introduced a diffusion-based segmentation framework that implicitly generated an ensemble of segmentation masks and proposed a novel metric, Collective Insight score, for assessing the performance of ambiguous models. More recently, Amit et al. \cite{amit2023annotator} introduced a novel DPM for binary segmentation that incorporated information from multiple annotations, creating a unified segmentation map reflecting consensus, which provided a unique approach to fuse multiple expert annotations.	

Despite the above achievements in 2D segmentation, DPM-based segmentation methods which enable accurate extraction of the organs and lesions from 3D data are needed for volumetric MRI. Diff-UNet \cite{Diff-UNet} was proposed for 3D multi-class segmentation with a label embedding operation converting the segmentation label maps into one-hot labels. During testing, it incorporated a step-uncertainty-based fusion module to fuse the multiple predictions during the denoising process to enhance the segmentation robustness. 

Fu et al. \cite{fu2023importance} improved 3D multi-class image segmentation with DDPM by tackling the issue of train-test inconsistency which caused degradation of the segmentation performance. Observing that the noise-corrupted ground-truth mask adopted during training may still contain morphological features, causing data leakage, the authors proposed a recycling training strategy to use the prediction from the previous steps instead of the noise-corrupted ground truth mask to predict the noise mask in the next step, aligning the training and inference process. In this work, the segmentation masks were directly predicted instead of sampled noise to facilitate the use of common segmentation loss of Dice loss and cross-entropy during training. Furthermore, Nichol et al. \cite{nichol2021improved} adopted a resampling variance scheduling to achieve a five-step denoising process for both training and inference, largely saving computation time and resources. To enhance the computational speed and storage efficiency for DPM-based 3D volume segmentation, Bieder et al. \cite{bieder2023diffusion} introduced PatchDDM, which was trained on coordinate-encoded patches, allowing for processing of large volumes in full resolution during sampling.

Akbar et al. \cite{akbar2023brain} explored the feasibility of using synthetic MRI data to train brain tumor segmentation models. They took four modalities of tumor MR images plus a one channel of tumor annotation into the DDPM and four GANs models to generate synthetic data with masks. The segmentation results indicated that the 2D-UNet segmentation model trained with synthetic images achieved similar performance metrics to that trained with real images. Compared with the existing GAN methods, the DDPM achieved competitive performance in synthesizing brain tumor images, while as the authors pointed out it is more likely to memorize the training images than GANs when the training dataset is too small. 

Different from supervised methods that require manual annotations during training, current weakly supervised and semi-supervised segmentation methods mainly consider DPMs as a semantic encoder of the image, and they realize pixel-wise segmentation by observing the differences of these semantic representations. For example, the study in \cite{baranchuk2021label} utilized a multi-layer perceptron to classify the latent representation of the DDPM, achieving the segmentation of natural images. The training of DPMs as representation learners does not rely on annotations, and the labeling supervision is mainly used to observe the differences between representations, which eliminates the requirement of laborious case-level annotation.

For MRI applications, Alshenoudy et al. \cite{alshenoudy2023semi} presented a semi-supervised brain tumor segmentation method under a scenario where annotated samples are scarce. This segmentation approach, developed from a method by \cite{baranchuk2021label}, adopted DDPM to learn visual representations of the input images for segmentation in an unsupervised way. However, the segmentation was achieved by fine-tuning the noise-predictor network on the labeled data instead of using a pixel-level classifier. Hu et al. \cite{hu2023conditional} innovatively explored conditional DPMs for locating the target objects by comparing the sampling under different conditions. Moreover, to amplify the difference caused by different conditions, this method extracted the semantic information from the gradient of the noise predicted by the DPM with respect to the condition. Experiments on different MRI datasets demonstrated its strong performance in brain tumor and kidney segmentation with only image-level annotations. The idea of observing the difference of representations in DPM-based segmentation can also be achieved by designing loss functions. Tursynbek et al. \cite{tursynbek2023unsupervised} designed a 3D generative DPM using a U-Net architecture as a feature extractor of 3D images for unsupervised segmentation. Unsupervised training with a composite loss enforcing feature consistency, visual consistency, and photometric invariance, the proposed method achieved superior segmentation performance in synthetic and real-acquired brain tumor MRI datasets. For future applications, in addition to exploring more ways of measuring the difference of representations, we expect more studies focusing on improving the efficiency and reliability of feature extraction using DPMs with segmentation-related priors. Additionally, application of the weakly-supervised segmentation DPMs to the challenging task of multi-label segmentation of high dimensional MR images is worth exploring.
\subsection{Anomaly Detection} \label{Anomaly-Detection}
Unlike DPM-based segmentation methods, which are mainly interested in generation masks, Anomaly detection aims to highlight anomalous regions by comparing the input image with the generated image containing healthy tissues. Therefore, DPMs with superior generation capability are becoming popular in anomaly detection tasks. Therefore, DPMs with superior generation capability are becoming popular in anomaly detection tasks. Table \ref{tab:Anomaly-Detection} summarizes the studies applying DPMs in anomaly detection of MR images, including the adopted DPM, the target organ, the learning strategy, the condition associated with the method, and the open-source code link. Specifically, we take whether the training of DPMs directly exploits each sample's known anomaly information as the learning strategy criteria.

\begin{table*}[h]
\centering
\fontsize{9}{10}\selectfont{

\begin{threeparttable}
    \caption{Emerging DPMs in MRI Anomaly Detection.}
    \label{tab:Anomaly-Detection}
\begin{tabular}{cccccccc}
\hline
 \multicolumn{1}{c}{Paper} & Method& \multicolumn{1}{c}{Organ}  & Dimension & Condition & Strategy &Training Data &\multicolumn{1}{c}{Code\tnote{*}} \\ \hline
  \cite{bercea2023mask} & DDPM & Brain & 2D & Mask Prior &Unsupervised & Healthy &\href{https://github.com/ci-ber/autoDDPM}{\textit{link}} \\
  \cite{iqbal2023unsupervised} & DDPM & Brain & 2D & Mask Prior& Unsupervised &Healthy &
 \href{https://github.com/hasan1292/mddpm}{\textit{link}}  \\
  \cite{behrendt2023patched} & DDPM & Brain & 2D & Patch Mask Prior& Unsupervised &Healthy &\href{https://github.com/FinnBehrendt/patched-Diffusion-Models-UAD}{\textit{link}}\\
  \cite{kascenas2023role} & DDPM & Brain & 2D\&3D &Noise Type&  Unsupervised &Healthy &  \href{https://github.com/antanaskascenas/denoisingae}{\textit{link}}\\
  \cite{pinaya2022fast} & DDPM & Brain & 2D & Latent Representation & Unsupervised &Healthy & - \\
  \cite{wolleb2022anomaly} & DDIM & Brain & 2D & Classifier Guidance & Supervised & Healthy + Unhealthy &\href{https://gitlab.com/cian.unibas.ch/diffusion-anomaly}{\textit{link}}\\
\cite{sanchez2022healthy} & DDPM & Brain & 2D & Classifier-Free Guidance &Supervised & Healthy + Unhealthy  & \href{https://github.com/vios-s/Diff-SCM}{\textit{link}}
\\ \hline
\end{tabular}
\begin{tablenotes}
    \item[*]"-" indicates the open-source code is not available.\\
\end{tablenotes}
\end{threeparttable}
}
\end{table*}

Wolleb et al. \cite{wolleb2022anomaly} first applied DPMs to anomaly detection in MRI. This work trained a DDPM and a binary classifier on datasets of healthy and diseased subjects. During inference, the input image was perturbed into a noisy image with the forward DDIM sampling, followed by the classifier-guided DDIM sampling process to generate images of healthy subjects. The anomaly detection was attained by calculating the anomaly map which is the pixel-wise difference between the generated and the original images. Sanchez et al. \cite{sanchez2022healthy} explored DPMs for brain lesion extraction. They found out that DPMs trained on only healthy data were insufficient to identify brain lesions. Then, they implemented a counterfactual DPM to generate healthy counterfactuals of given input images with implicit guidance, attention-based conditioning, and dynamic normalization to enable the localization of brain lesions. Anomaly detection was subsequently achieved by comparing the factual input and counterfactual output images. Pinaya et al. \cite{pinaya2022fast} proposed an unsupervised anomaly detection method that adopted VQ-VAE and DDPM. The VQ-VAE was used to obtain the latent representation of an input image. Then the DDPM learned the distribution of the latent representation of healthy data. During inference, the KL-divergence was calculated to evaluate the proximity of each reverse step to the expected Gaussian transition, so this method can obtain the mask of the anomalies by thresholding the KL Divergence. The marjor motivation of this setting is that if the input image is from a healthy subject, the reverse step only removes the added Gaussian noise; if the image contains anomalies, each reverse step will also remove parts of the anomalous region's signal, leading to a high KL Divergence. The anomaly mask was then used in the reverse process to correct the anomalies in the latent space, on which the VQ-VAE decoder was performed to obtain the output image with anomalies corrected. 

In order to investigate the role of noise in denoising models based abnormalities detection, the study in \cite{kascenas2023role} compared three types of noise (Gaussian, Simplex or coarse) for a classical denoising autoencoder and the DPM-based method \cite{pinaya2022fast, wyatt2022anoddpm} on the 2D head MRI and 3D head CT dataset, respectively. The results indicated that noise type indeed impacted the performance of denoising models for anomaly detection, and the coarse noise outperformed the other two noise types. Regarding the denoising models, the authors found that the simple denoising autoencoder with optimal noise performed better than the more advanced DPMs, while DPMs demonstrated the capability of "healing" anomalies and generating convincing high-definition reconstructions.

The previously mentioned DPMs for anomaly detection performed noise estimation across the entire image. Behrendt et al. \cite{behrendt2023patched} argued that performing noise estimation on the whole image makes it difficult to accurately reconstruct the complex structure of the brain. Therefore, they applied a patch-based DDPM proposed in \cite{ozdenizci2023restoring} to generate image patches which were stitched together to obtain the final healthy brain MR images for calculating the anomaly score.

Furthermore, Iqbal et al. \cite{iqbal2023unsupervised} presented a method called masked-DDPM, which added masking-based regularization by masking the input image in the spatial image domain and frequency domain before inputting to the DDPM for training. The masking strategy imposed a constraint on DDPM for generating healthy images during inference regardless of the input images. To enhance the generalization ability of DPMs in detecting diverse types of anomalies, AutoDDPM \cite{bercea2023mask} integrated the masking, stitching, and resampling operations. Specifically, the pre-trained DDPM generated pseudo-healthy samples under the automatic mask setting, which were then stitched to the unmasked original healthy tissues in the denoising process. Subsequently, resampling of the joint noised distributions achieved harmonization and in-painting effects, generating good-quality pseudo-healthy reconstructions.

\subsection{Further Research Topics}\label{Other-Applications}

Although DPMs have proved to be a useful tool in the aforementioned various MRI tasks, there are still other issues in MRI that can be addressed by DPMs, which only have some preliminary studies. Table \ref{tab:Further-Research-Topics} summarizes the topics, the adopted DPM, the target organ, the highlights of each study and the open-source code link. In the following, we will briefly introduce these pioneering studies from two perspectives: Imaging Enhancement and Other Tasks.

\begin{table*}[h]
\centering
\fontsize{9}{10}\selectfont{
\begin{threeparttable}
    \caption{Emerging DPMs in further research topics in MRI.}
    \label{tab:Further-Research-Topics}
\begin{tabular}{cccccc}
\hline
Task & \multicolumn{1}{c}{Paper} & \multicolumn{1}{c}{Method} & \multicolumn{1}{c}{Organ} & \multicolumn{1}{c}{Key Points} & \multicolumn{1}{c}{Code\tnote{*}} \\ \hline

 \multirow{-1}{*}{Denoising}& \cite{xiang2023ddm} & DDPM & Brain; Knee & Self-Supervised Denoising & {\href{https://github.com/StanfordMIMI/DDM2}{\textit{link}}}\\

\midrule

\multirow{-1}{*}{Image Registration} & \cite{kim2022diffusemorph} & DDPM & Cardiac; Brain & 2D \& 3D Conditional Registration  & -\\

\midrule
& \cite{levac2023accelerated} & SMLD & Brain & Reconstruction and Motion Correction& \href{https://github.com/utcsilab/motion_score_mri}{\textit{link}}\\
\multirow{-2}{*}{Motion Correction} & \cite{oh2023annealed} &  SDE  & Brain; Liver & Motion Correction with k-space Consistency& - \\

\midrule
 
 & \cite{wu2023super} & DDPM & Brain & 2D Image Super-Resolution& - \\
 & \cite{mao2023disc} & DDIM & Brain & Multi-contrast Image 2D Super-Resolution& \href{https://github.com/yebulabula/disc-diff}{\textit{link}} \\
 \multirow{-3}{*}{Super Resolution} & \cite{chung2022mr} & SDE & Liver & 2D k-space Denoising \& Super-Resolution & - \\

\midrule
 & \cite{chen2023seeing} & LDM & Brain &  Generating visual images from fMRI& \href{https://github.com/zjc062/mind-vis}{\textit{link}}\\
 \multirow{-2}{*}{Sematic Understanding}& \cite{takagi2023high} & LDM & Brain & Generating visual images from fMRI& \href{https://sites.google.com/view/stablediffusion-with-brain/}{\textit{link}}\\

\midrule
\multirow{-1}{*}{Inpainting}& \cite{rouzrokh2022multitask} & DDPM & Brain & Image Inpainting & \href{https://github.com/Mayo-Radiology-Informatics-Lab/MBTI}{\textit{link}}\\
\midrule
\multirow{-1}{*}{Classification} & 
\cite{ijishakin2023interpretable} & DDIM & Brain & Alzheimer's Disease Classification  & -\\
\hline
\end{tabular}

\begin{tablenotes}
    \item[*]"-" indicates the open-source code is not available.
\end{tablenotes}
\end{threeparttable}
}
\end{table*}

\subsubsection{Imaging Enhancement}

As an emerging paradigm for generating images using noise estimation, one of the most direct applications of DPMs is removing random noise introduced during imaging. Xiang et al. \cite{xiang2023ddm} designed a self-supervised denoising method based on DDPM for both 2D and 3D diffusion-weighted MRI. However, recent research has revealed the significant potential of DPMs for more specialized topics in imaging enhancement, such as motion correction, super-resolution and semantic understanding.

\paragraph{Motion Correction} 
Motion artifact reduction is an active research area in MRI, for which numerous deep learning methods have been developed. However, most of these deep learning methods require paired motion-free and motion-corrupted images for supervised training, which are difficult to obtain in practice. The model trained with simulated images with motion artifacts may not generalize will to real motion artifacts. To address this issue, Levac et al. \cite{levac2023accelerated} proposed a method to simultaneously reconstruct undersampled MR images and estimate rigid head motion using a score-based DPM. While the score-based DPM was supervised with simulated motion data, it was agnostic to the forward model including the sampling mask and the motion pattern, making it applicable to real MR acquisitions with unpredictable patient movements. Recently, Oh et al. \cite{oh2023annealed} proposed an annealed score-based method for respiratory motion artifacts reduction in abdominal MR images. The DPM trained on motion-free images was able to removed motion artifacts by using a repetitive diffusion-reverse process and adding low-frequency consistency in each step of the reverse process. 

\paragraph{Super-Resolution} High-resolution MRI images are beneficial for delineating fine anatomical structures and small lesions. However, acquiring high-resolution images is challenging due to limitations such as magnetic field strength, signal-to-noise ratio, and acquisition time. Super-resolution aims to recover high-frequency information for low-resolution inputs. The adoption of DPMs for MRI super-resolution can be in the image or the acquired k-space domain. In the image domain, the low-resolution image typically serves as a condition of generating the high-resolution image \cite{wu2023super}. Moreover, for multi-contrast MRI, Mao et al. \cite{mao2023disc} proposed a framework combining a disentangled U-Net backbone with the guided-DDIM \cite{dhariwal2021diffusion} that could leverage the complementary information between contrasts for super-resolution. In the k-space domain, Chung et al. \cite{chung2022mr} proposed a score-based SDE to generate the high-frequency components, while the low-frequency signals were preserved in a regularization manner.

\paragraph{Semantic Understanding} As a specific application of MRI that can reflect the brain activity, functional magnetic resonance imaging (fMRI) contains a wealth of information related to visual functions. There are studies exploring whether DPMs can be utilized to explore the visual semantic information embedded in fMRI data, or even directly recover visual images. Chen et al. \cite{chen2023seeing} developed the MinD-Vis model with two main stages to address the challenge of reconstructing high-quality images with correct semantic information from fMRI signals. Inspired by the sparse coding of information in the primary visual cortex, the first stage of their model represented fMRI data as a sparsely-encoded representation with local constraints. Then, the visual content was generated with the encoded representation in the second stage using a double-conditioned LDM and end-to-end fine tuning. Takagi et al. \cite{takagi2023high} combined three developments in their earlier work: decoded text from brain activity, nonlinear optimization with GAN for structural image reconstruction, and decoded depth information from brain activity with an LDM, to generate images with accurate semantic information.

\subsubsection{Other Tasks}

\paragraph{Image Registration} Registration algorithms using generative learning have shown to be effective in aligning different MRI scans. The fundamental idea is to use a network to obtain a deformation field between the moving and fixed images which is then used to warp the moving image to achieve registration. Kim et al. \cite{kim2022diffusemorph} first reported a deformation framework for 2D facial expression and 3D cardiac MRI registration using DPMs. One part of this framework is a diffusion network that learns a conditional score of the motion field between the moving and fixed images. Another part of this framework is a deformation network that can utilize the learned conditional score to estimate the deformation field and produce deformed images. Notably, the learned latent feature of the diffusion network contained spatial information, which can then be linearly scaled to generate motion fields along a continuous trajectory from the fixed to the moving images.

\paragraph{Inpainting} For inpainting, Rouzrokh et al. \cite{rouzrokh2022multitask} constructed a 2D axial slice inpainting tool using DDPM that can add high-grade glioma and the corresponding tumor components or normal brain tissue in user-specified regions, which could address the problem of insufficient high-grade glioma data in practice.

\paragraph{Classification} For classification, Ijishakin et al. \cite{ijishakin2023interpretable} proposed to utilize the cosine similarity between the latent codes of DDIM and the hidden variable of the category semantic encoder to classify Alzheimer's Disease. This method achieved comparable classification performance to black-box models while improved model interpretability.

\section{Trends and Challenges}\label{Sec:Trends-and-Challenges}

Accompanied by the rapid development of the methodologies of DPMs and the increasing attention to the application of large generative models, DPMs have shown strong potential for application in different MRI tasks. In MRI, it is desirable to have high-resolution, artifacts-free, and multi-contrast images for accurate diagnosis. DPM as an effective method of generating high-fidelity samples has achieved remarkable performance in MR image reconstruction, which has drawn more attention than other tasks as shown in Fig. \ref{fig:Sec1-PieHist}(b). Through the two processes of adding noise to data and removing noise to reach the desired data distribution, DPMs are able to capture the complex relationships between signals and noise/artifacts. Chung et al. \cite{chung2022score} demonstrated that DPM with score-based SDE trained with magnitude-only images could generalize to single-coil and multi-coil complex data, and was also robust to different under-sampling patterns, which seems impossible for previous non-DPM methods. Additionally, DPMs are becoming popular in MR image translation and generation due to their powerful capability of generating images with good quality and high diversity conditionally and unconditionally. Furthermore, it is also observed that DPM can serve as an effective representation learner for discriminative tasks. Since there is no need to learn additional encoders to map images to latent spaces, DPMs enjoy distinctive advantages in segmentation tasks.

While DPMs have demonstrated great potential in several MRI tasks, by analyzing the reviewed studies, we identify specific trends and challenges of applying DPMs in MRI. In the following, we share our opinions about research directions on model designs and expanding applications.

\subsection{Model Design}

\paragraph{Accelerated Sampling}
One of the main characteristics of diffusion probabilistic models is the requirement of a large number of steps to obtain high-quality samples. Therefore, the exploration of efficient sampling methods to improve the generation speed is advantageous for the widespread application of DPMs in MRI. Yang et al. \cite{yang2022diffusion} summarized two mainstream approaches for sampling acceleration in DPMs: learning-free sampling and learning-based sampling. Learning-free sampling represents a type of method for achieving accelerated sampling without the need for additional learning. For instance, Wizadwongsa et al. \cite{wizadwongsa2023accelerating} provided a solution based on operator splitting methods to reduce the sampling time, and Lu et al. \cite{lu2022dpm} solved the diffusion ODE with the data prediction model to reduce the step size. Chung et al. \cite{chung2023fast} proposed to decompose the intermediate sampling result into two orthogonal parts of clean and noise data manifolds and utilized conjugate gradient update in data consistency to ensure that the intermediate reconstruction falls on the clean manifold, achieving more accurate and faster reconstruction. Learning-based sampling refers to those methods that require the learning of a solver beyond the training of DPMs. For example, Chung et al. \cite{chung2022come} proposed to start the reverse sampling process with a better initialization such as the prediction of some pre-trained neural network instead of a random noise, which can significantly reduce the number of sampling steps. Similarly, Zheng et al. \cite{zheng2022truncated} designed an adversarial auto-encoder to learn an implicit distribution to start the reverse process. Luhman et al. \cite{luhman2021knowledge} proposed an accelerated method for image generation using knowledge distillation.

\paragraph{Application for High-dimensional MRI} DPMs have achieved remarkable performance in MRI reconstruction, denoising and super-resolution. However, most of the works train DPMs in the pixel space, where the variable at each diffusion time step shares the same dimension to the original data. For high-dimensional MRI data with extra contrast or temporal dimensions, if processed separately by DPMs, the inter-contrast or temporal correlation cannot be exploited. If learned simultaneously, the computation burden may be increased significantly as DPMs do not reduce data dimensions. LDMs \cite{rombach2022high}, which work in a much lower-dimensional latent space instead of the pixel space may provide a viable solution. In LDMs, the variational auto-encoder is leveraged, where an encoder compresses the data into a latent space, and then DPMs are applied in the latent space, after which, a decoder maps the diffusion generations from latent to data space. However, applying LDMs to MRI reconstruction may be challenging, as it is difficult to guarantee data consistency in the latent space. Song et al. \cite{song2023solving} recently proposed an algorithm that enforced data consistency by solving an optimization problem during the reverse sampling process, after which a novel resampling scheme was designed to map the measurement-consistent sample back onto the correct data manifold. This method worked well for solving both linear and non-linear inverse problems, and provided a promising paradigm for applying DPMs in high-dimensional MRI.

\paragraph{Incorporating Prior} Incorporating MRI prior into the noise estimation and sampling of DPMs is a common way to reduce the randomness in generating MRI data. Specifically, during sampling of the reverse process, prior information such as observation patterns \cite{xie2022measurement} and mask labels \cite{dorjsembe2023conditional} are usually added through data consistency constraints. Another approach incorporated prior information into the learning parameters, such as particular scoring designs \cite{aali2023solving} and conditional generation based on measurement modality \cite{song2021solving}. MRI offers abundant physical priors that can be used to guide model training. Recent works such as \cite{cui2023spirit} and \cite{peng2023one} have already started to look into the incorporation of MRI physics model into the design of DPMs. Designing DPMs that incorporate relevant MRI priors represents a promising direction for improving the generation quality of DPMs.

\begin{table*}[htp]
\centering
\fontsize{9}{10}\selectfont{
    \begin{threeparttable}
    \caption{Comparisons of DPMs with other deep learning methods in reconstruction, segmentation, and anomaly detection tasks.}
    \label{tab:comparison}
\begin{tabular}{cccccc}
\hline
Task & Experimental Setup & Paper & Method & \multicolumn{2}{c}{Metric} \\
\hline
\multirow{13}{*}{Reconstruction} & \multirow{5}{*}{\makecell[c]{Reconstruction on the fastMRI\cite{knoll2020fastmri}\\acceleration factor = 4\\knee PD dataset}} &  &  & PSNR & SSIM \\\cline{5-6}
 &  & - & zero-filled reconstruction & 29.62\tnote{a} & 0.745\tnote{a} \\
 &  & \cite{ronneberger2015u} & CNN & 34.04\tnote{a} & 0.834\tnote{a} \\
 &  & \cite{xie2022measurement} & DDPM & 36.69 & 0.905 \\
 &  & \cite{chung2022come} & SDE & 32.51 & - \\\cline{2-6}
 & \multirow{9}{*}{\makecell[c]{Reconstruction on the fastMRI\cite{knoll2020fastmri}\\acceleration factor = 4\\brain T1 dataset}} &  &  & PSNR & SSIM \\\cline{5-6}
 &  &\cite{aggarwal2018modl} & CNN & 39.8\tnote{b} & 0.893\tnote{b} \\
 &  & \cite{narnhofer2019inverse} & GAN & 31.7\tnote{b} & 0.746\tnote{b} \\
 &  & \cite{gungor2023adaptive} & DDPM & 40.2 & 0.959 \\
 &  & \cite{peng2022towards} & DDPM & 40.2 & 0.953\tnote{c} \\
 &  & \cite{peng2022towards} & DDPM(Self Supervised) & 38.4\tnote{c} & 0.958\tnote{c} \\
 &  & \cite{dar2020prior} & RGAN & 39.6\tnote{c} & 0.955\tnote{c} \\
 &  & \cite{dar2020prior} & RGAN(Self Supervised) & 38.3\tnote{c} & 0.950\tnote{c} \\
 &  & \cite{korkmaz2023self} & DDPM & 40.1 & 0.965\\
 \hline
\multirow{7}{*}{Segmentation} & \multirow{7}{*}{\makecell[c]{3D Segmentation \\on the BraTS2020 \cite{menze2014multimodal,   bakas2018identifying, bakas2017advancing}}} &  &  & Dice & HD95 \\ \cline{5-6} 
 &  & \cite{hsu2021brain} & CNN & 82.73\tnote{d} & 3.494\tnote{d} \\
 &  & \cite{oktay2018attention} & CNN & 78.09\tnote{d} & 13.549\tnote{d} \\
 &  & \cite{hatamizadeh2021swin} & Transformer & 83.04\tnote{d} & 3.891\tnote{d} \\
 &  & \cite{hatamizadeh2022unetr} & Transformer & 81.55\tnote{d} & 4.896\tnote{d} \\
 &  & \cite{Diff-UNet} & DDIM & 85.35 & 3.389 \\
 &  & \cite{bieder2023diffusion} & DDPM & 88.80 & 9.040 \\ \hline
\multirow{12}{*}{Anomaly   Detection} & \multirow{12}{*}{\makecell[c]{2D Anomaly Detection\\ on the BraTS2021\cite{baid2021rsna}}} &  &  & \multicolumn{2}{c}{Ideal Dice} \\ \cline{5-6} 
 &  & \cite{meissen2021challenging} & Thresholding & \multicolumn{2}{c}{0.667\tnote{e}} \\
 &  & \cite{schlegl2019f} & GAN & \multicolumn{2}{c}{0.316\tnote{e}} \\
 &  & \cite{zimmerer2019unsupervised,zimmerer2018context} & VAE(reconstruction) & \multicolumn{2}{c}{0.395\tnote{e}} \\
 &  & \cite{chen2020unsupervised} & VAE(restoration) & \multicolumn{2}{c}{0.685\tnote{e}} \\
 &  & \cite{kascenas2022denoising} & DAE & \multicolumn{2}{c}{0.758\tnote{e}} \\
 &  & \cite{iqbal2023unsupervised} & DDPM & \multicolumn{2}{c}{0.530} \\
 &  & \cite{behrendt2023patched} & DDPM & \multicolumn{2}{c}{0.490} \\
 &  & \cite{kascenas2023role} & DDPM & \multicolumn{2}{c}{0.773\tnote{f}} \\
 &  & \cite{pinaya2022fast} & DDPM & \multicolumn{2}{c}{0.537} \\
 &  & \cite{wolleb2022anomaly} & DDIM & \multicolumn{2}{c}{0.745\tnote{e}} \\
 &  & \cite{sanchez2022healthy} & DDPM & \multicolumn{2}{c}{0.762} \\ \hline
\end{tabular}
\begin{tablenotes}
\item[a] This result is reproted by Xie et al.\cite{xie2022measurement}\quad \quad \quad \quad
\item[b] This resultis reported by G{\"u}ng{\"o}r et al. \cite{gungor2023adaptive}\\
\item[c] This result is reported by Korkmaz et al. \cite{korkmaz2023self}\ \  \ \
\item[d] This result is reported by Xing et al.\cite{Diff-UNet}\\
\item[e] This result is reported by Sanchez P et al.\cite{sanchez2022healthy}\quad  
\item[f] Training data is selected.
\end{tablenotes}
\end{threeparttable}
}
\end{table*}

\paragraph{More Comparative Studies}

DPMs have demonstrated their potential in methodological innovation among various MRI tasks. The evaluation results between DPMs and other well-known methods under the comparable setting in two reconstruction studies, one segmentation, and one anomaly detection experiments, as in Tabel \ref{tab:comparison}, have indicated outstanding performance of DPMs in these MRI tasks. However, we note that the application of DPMs in MRI is still at early stage, the comparative results reported in public datasets may not be gathered for all relevant tasks. For wide applications of DPMs in MRI, their fair comparisons with other deep learning methods are warranted for various MRI tasks.

\subsection{Expanding Applications}
\begin{table*}[h]
\fontsize{9}{10}\selectfont{
\begin{threeparttable}

\centering
    \caption{Public MRI datasets commonly used in DPMs.}
    \label{tab:Public-MRI-Dataset}
\begin{tabular}{cccccc}
\hline
Dataset & Paper & Organ & Dataset & Paper & Organ\\ \hline
3D Stanford& \cite{epperson2013creation} & Knee & GOD & \cite{horikawa2017generic} & Brain \\
ABIDE & \cite{di2014autism} & Brain & Gold Atlas & \cite{nyholm2018mr} & Pelvic \\
ACDC & \cite{bernard2018deep} & Cardiac & HCP & \cite{van2013wu} & Brain \\
ADNI & \cite{crawford2016image} & Brain & IXI & \cite{IXI} & Brain \\
AMOS & \cite{ji2022amos} & Spleen, Kidney, Gallbladder, etc.& MRNet  & \cite{bien2018deep} & Knee \\
AOMIC & \cite{snoek2021amsterdam} & Brain & MSD & \cite{antonelli2022medical} &  Brain, Cardiac, Lung, etc.\\
ATLAS V2.0 & \cite{liew2022large} & Brain & MS-MRI & \cite{carass2017longitudinal} & Brain \\
BOLD5000 & \cite{chang2019bold5000} & Brain & NSD & \cite{allen2022massive} & Brain \\
BrainAge & \cite{feng2020estimating} & Brain & NTUH & \cite{wu2021deep} & Brain \\
BraTS2018 & \cite{menze2014multimodal} & Brain & OASIS & \cite{marcus2007open} & Brain \\ 
BraTS2019 & \cite{menze2014multimodal, bakas2018identifying, bakas2017advancing} & Brain & OASIS-3 & \cite{lamontagne2019oasis} & Brain \\
BraTS2020 & \cite{menze2014multimodal, bakas2018identifying, bakas2017advancing} & Brain &  PICAI  & \cite{saha2023artificial} & Prostate \\
BraTS2021 & \cite{baid2021rsna} & Brain & QUBIQ & \cite{QUBIQ} & Brain, Prostate, Kidney \\
CHAOS & \cite{kavur2021chaos} & Kidney & SABRE & \cite{jones2020cohort} & Brain, Cardiac \\
CMPS & \cite{li2022prototypical} & Prostate & SKM-TEA & \cite{desai2021skm} & Knee \\
CuRIOUS & \cite{xiao2019evaluation} & Brain & SRI-Multi & \cite{zhang2022multi} & Brain\\
DUKEBrest & \cite{clark2013cancer} & Brest & UCSF-PDGM & \cite{calabrese2022university} & Brain \\
fastMRI & \cite{knoll2020fastmri} & Knee, Brain & UKB & \cite{sudlow2015uk} & Brain \\
fastMRI+ & \cite{zhao2021fastmri+} & Knee, Brain &  WMH & \cite{kuijf2019standardized}& Brain \\
\hline
\end{tabular}
\end{threeparttable}
}
\end{table*}

\begin{figure}[h]
	\centering
        \includegraphics[width=0.95\linewidth]{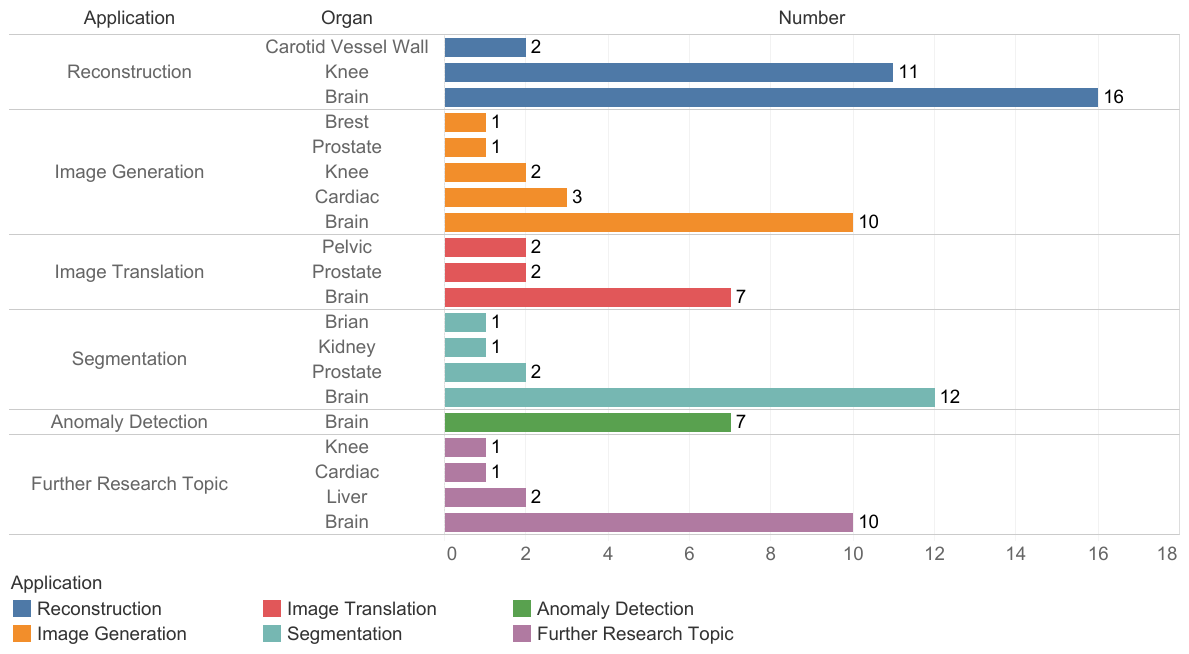}
	\caption{Organs considered by DPMs in different MRI applications.}
 \label{fig:sec4OrganTask}
\end{figure}

\paragraph{Organ \& Tasks} DPMs have demonstrated powerful capabilities for accurately portraying data distributions and controllably generating high-quality samples. However, training a DPM with these capabilities usually requires a large quantity of high-quality MRI samples. Since acquiring MRI data is relatively expensive, data abundance remains one of the significant challenges for applying DPMs in MRI. Obviously, the data availability in different MRI application scenarios has a direct influence on the organs that DPMs focus on. From the summarized organs in the tables of different applications of DPMs in MRI, as also shown in Fig. \ref{fig:sec4OrganTask}, it can be seen that the number of studies focusing on brain largely surpasses other organs, which is because there is a wealth of public datasets of brain MRI. The public MRI datasets that have been adopted in DPMs are summarized in Table \ref{tab:Public-MRI-Dataset}. In comparison, applications of DPMs in the thoracic and abdominal regions such as the heart, kidneys, and prostate have been less frequently reported. Possible reasons are that the available datasets of these body parts are scarce and that some applications related to these regions are more challenging which may require further development of DPMs. 

Though less investigated, thoracic and abdominal MRI hold great potential for DPMs due to the unique physiological features and acquisition challenges. For example, to mitigate respiratory and cardiac motion, the coverage and spatial resolution of acquired cardiac images are usually compromised for a reasonable scan duration, where DPMs can be used to enhance the reconstruction quality and resolution of cardiac images. Furthermore, there tend to be motion artifacts in the abdominal and cardiac MRI images. A fundamental challenge of previous deep-learning based motion artifact reduction methods is the requirement of paired motion-free and motion corrupted images for model training which can be difficult to acquire or simulate. One of the primary strengths of DPMs is the ability to work without paired label data. The work in \cite{oh2023annealed} demonstrated that a score-based method trained with only motion-free images can effectively reduce motion artifacts during reverse diffusion process, and outperformed the GAN-based method. Thus, the potential of DPMs for MR motion artifacts reduction is worth further exploring. All in all, we note that high-quality and diverse publicly available MRI datasets are in demand to facilitate the exploration of DPMs in more MRI tasks.

\paragraph{Privacy Protection} Although DPMs have demonstrated superiority over other generative models in many application scenarios, it is essential to acknowledge that due to the setting of the reverse process of generating samples that follow the distribution of training data, there may be an increased risk of patient privacy leakage from DPMs compared to other generative models \cite{carlini2023extracting}. 

Therefore, protecting patient privacy during the training and application phases becomes crucial in utilizing DPMs for wide clinical applications. There are already some emerging solutions in natural images. In the training phase, Dockhorn et al. \cite{dockhorn2022differentially} proposed a method that combined rigorous differential privacy into the training of DPMs to ensure that the generated results cannot be judged whether they come from the training data. Moreover, \cite{liu2023diffprotect} incorporated adversarial semantic code into the DDIM and applied semantic regularization to add imperceptible semantic perturbation to the final images, which can protect the identity privacy implicitly in the training data. For the application phase, the combination of DPMs with federated learning has also sprung up with works such as \cite{jothiraj2023phoenix} and \cite{yang2023exploring}, which not only alleviates the problem of burdensome computation of DPMs but also makes it possible to apply DPMs more privacy-friendly. All such works provide promising solutions of enhancing the protection of patient privacy contained in MRI data.

\paragraph{Trustworthy DPMs} DPMs and their applications to medical imaging, including MRI, are still at the early stage. The vast majority of existing studies of DPMs are focused on technique developments. Only a few studies involved clinical assessments in DPM evaluation. Khader et al. \cite{khader2022medical} utilized quantitative assessment from two experienced radiologists in evaluating the synthetic capability of the developed DPM model. Saeed et al. \cite{saeed2023bi} evaluated the generated pathology by a urological MR expert. Further studies on the clinical implementation of DPM-based methods in different MRI tasks are of great value. In the context of trustworthy AI, constructing a trustworthy DPM in MRI is essential for its clinical adoption, the key of which lies in the stability and reliability of the generated results. In addition to designing evaluation metrics that comply with clinical requirements and uncertainty measures of generated results as priors or conditional guidance for DPMs, research on adversarial attacks on the backbone of DPMs can provide new ideas for building robust and trustworthy DPMs in MRI. Current works such as \cite{nie2022diffusion} and \cite{teneggi2023trust} have investigated adversarial attacks on DPMs in natural images and even in medical images. Aiming for clinical applications, we envision that there will be more studies in the near future working on the construction of trustworthy DPMs in medical imaging.
\section{Conclusion}\label{Sec:Conclusion}
In this paper, we provided a comprehensive and timely review focusing on the application of DPMs in various MRI tasks. With the aim of helping researchers to understand the fundamentals of DPMs, we first introduced two dominant classes of DPMs from a unified view of the diffusion and reverse processes and summarized the current development of conditional DPMs. Next, we reviewed studies applying DPMs in various MRI tasks, including reconstruction, image generation and translation, segmentation, anomaly detection, and other pioneering research topics, aiming to help researchers to find the latest literature according to the practical needs. For each task, in addition to the introduction of related work, we also provided a table containing the adopted DPM, target organ, task-related key points, and available open-source code link, which can offer useful references and sources to researchers who are interested in adopting DPMs. Furthermore, we pointed out challenges and future directions of applying DPMs in MRI, where researchers can benefit from practical insights on model design, data resources, and evaluation methods. Since DPMs in MRI are growing rapidly, this review may not cover all the studies. However, we spared no effort to gather relevant and high-quality papers. We believe that the insights presented in this paper about DPMs in MRI may serve as a good reference for researchers interested in this field and contribute to further advancements.

\bibliographystyle{unsrt}
\bibliography{main}
\end{document}